\documentclass[letterpaper]{article}
\usepackage[letterpaper]{geometry}

\usepackage[utf8]{inputenc}
\usepackage[T1]{fontenc} 
\usepackage[cjk]{kotex}
\usepackage{amsmath}
\usepackage{graphicx}
\usepackage{url}
\usepackage[table]{xcolor}

\usepackage{latexsym}

\usepackage{hyperref}
\definecolor{darkblue}{rgb}{0, 0, 0.5}
\hypersetup{colorlinks=true,citecolor=darkblue, linkcolor=darkblue, urlcolor=darkblue}

\usepackage{tabularx}
\usepackage{booktabs}

\usepackage{setspace}

\usepackage[authoryear,round]{natbib}
\usepackage{algorithmic,algorithm}
\renewcommand{\algorithmiccomment}[1]{\bgroup\hfill$\triangleright$~#1\egroup}

\usepackage[linguistics]{forest} 
\usepackage{synttree}
\usepackage{subcaption}
\usepackage{tikz-dependency}

\usepackage{CJKutf8}

\usepackage{gb4e}

\title{\textbf{
{Representing and Parsing Korean Constituency Structure at Different Levels of Granularity}\thanks{Corresponding author: \url{jungyeul@kaist.ac.kr}}
}}
\author{
Jungyeul Park$^{1}$, KyungTae Lim$^{1}$, \\
Zihao Huang$^{2}$, Eunkyul Leah Jo$^{2}$, Yige Chen$^{3}$, Chulwoo Park$^{4}$\\
$^{1}$Korea Advanced Institute of Science \& Technology, South Korea\\
$^{2}$The University of British Columbia, Canada\\
$^{3}$The Chinese University of Hong Kong, Hong Kong\\
$^{4}$Anyang University, South Korea
}
\date{ 
}

\begin{document}
\maketitle
\begin{abstract}
Korean constituency parsing raises a representational challenge because the terminal units of a phrase-structure tree do not straightforwardly correspond to simple surface words. Korean eojeols are morphologically complex spacing units, and existing constituency resources differ in how they represent eojeol-internal morphology and non-overt elements. This paper compares three constituency parsing representations derived from the Penn Korean Treebank: Morpheme+XPOS, Eojeol+XPOS, and Eojeol+UPOS. We construct these representations by removing null elements, aligning Penn Korean phrase structure with overt eojeol tokens, preserving Penn Korean phrase labels where possible, and varying the terminal and preterminal layers. We then evaluate canonical non-binary transition-based constituency parsers in top-down, in-order, and bottom-up orders under a shared modeling and evaluation setup. All experiments use gold terminal segmentation and gold preterminal labels and therefore evaluate constituency parsing conditioned on gold morphosyntactic annotation. Eojeol terminals yield shorter transition sequences, but Eojeol+UPOS parsing substantially underperforms the morphologically richer conditions. Eojeol+XPOS narrows this gap, while Morpheme+XPOS gives the strongest results even after its predictions are projected to the eojeol terminal domain. Under these gold-annotation conditions, the results show that fine-grained morphological and XPOS representations provide valuable evidence for the evaluated parsers. This empirical finding concerns the information available for parsing and does not by itself determine the linguistically preferable terminal domain. Independently, linguistic and resource-design considerations motivate eojeol as a stable and interpretable surface domain for phrase-structure annotation, with morpheme-level and XPOS information retained as aligned morphosyntactic evidence.

\end{abstract}

\doublespacing

\section{Introduction}

{Korean constituency parsing poses a representational question that is less apparent in languages such as English or Chinese: What should count as the terminal unit in a phrase-structure tree?} Korean {eojeols} are morphologically complex spacing units, typically combining lexical stems with case particles, verbal endings, derivational suffixes, auxiliaries, or other functional morphemes. At the same time, {eojeols} are the surface units over which Korean written sentences are segmented, and they provide the token domain used by most Korean dependency resources. Constituency parsing for Korean therefore requires more than a phrase-label inventory or a parsing algorithm. It requires an explicit decision about how phrase structure, surface wordhood, and morphology should be related.

This paper investigates eojeol as a stable and linguistically interpretable terminal domain for Korean constituency representation and annotation. We also consider a separation between the constituency domain and the information used to recover it, with morpheme-level information retained as valuable evidence for parsing and aligned morphosyntactic annotation.
The claim is not that Korean syntax is independent of morphology. Functional morphology is central to Korean syntax: case particles and topic markers signal nominal functions, connective and adnominal endings mark clausal relations, and verbal endings encode tense, mood, modality, clause type, and subordination. The question is whether these bound morphemes should be projected as independent constituency terminals. Following the synthetic view of Korean wordhood, in which functional morphemes are grammatically integrated within a larger syntactic unit rather than functioning as independent syntactic words \citep{chulwoo-2019-for-developing}, we treat {eojeols} as the terminal units of constituency structure, while preserving morpheme sequences and fine-grained Korean {part-of-speech (POS)} tags as aligned morphosyntactic annotation \citep{park-park-2026-constituency}.

This separation is important for both linguistic interpretation and resource design. If morphemes are used as constituency terminals, word-internal morphological composition is projected into the phrase-structure yield. Such a representation makes morphosyntactic information directly visible, but it also intertwines morphology-internal segmentation with phrase-level grouping. Bound functional material may then appear in the same representational space as phrasal constituents, and constituency boundaries may partly reflect morphological attachment rather than surface phrase structure. An eojeol-based representation assigns different roles to the two layers: phrase structure is defined over overt spacing units, while morphology remains available as an auxiliary layer for analysis, parsing, and downstream interpretation.

The issue is especially relevant because Korean syntactic resources have developed under different assumptions about morphology and terminal granularity. Treebanks constitute an essential resource for natural language processing \citep{abeille:2003}, and several Korean phrase-structure resources have been created, including the KAIST treebank \citep{choi-EtAl:1994}, the Penn Korean Treebank \citep{han-EtAl:2002}, and the {Sejong treebank}. These resources differ not only in phrase labels and structural conventions, but also in how they relate phrase structure to Korean morphology. KAIST-style annotation is closely tied to morpheme-level analysis, Sejong-style phrase structure has commonly been distributed in binarized form, and the Penn Korean Treebank combines Penn-style phrase-structural annotation with terminal-internal Korean morphological analysis. The Penn Korean Treebank also includes explicit null elements for empty arguments, traces, empty operators, and predicate deletions \citep{han-han-ko:2001}. These differences make Korean treebanks linguistically rich, but they also make it difficult to compare parsing results unless the terminal domain is made explicit.

The Penn Korean Treebank provides a controlled setting for studying this issue because its overt surface units can be aligned with eojeol-level tokens and internal morphological annotation. We convert the treebank by removing null elements from the parsing target, recovering overt eojeol terminals, preserving Penn Korean phrase labels where possible, and aligning each eojeol with its morpheme sequence, language-specific detailed XPOS annotation, and universal POS (UPOS) label \citep{petrov-das-mcdonald:2012:LREC}. From this aligned representation, we derive three parsing targets: a morpheme-based representation with XPOS preterminals, an eojeol-based representation with XPOS preterminals, and an eojeol-based representation with UPOS preterminals.

These three targets separate two dimensions that are often conflated. The first is terminal granularity: whether constituency structure is defined over morphemes or over {eojeols}. The second is morphosyntactic category granularity: whether terminal categories are represented by fine-grained Korean XPOS labels or by coarser UPOS labels. In the Morpheme+XPOS representation, particles, endings, derivational suffixes, and other morphemes appear as independent terminals. In the Eojeol+XPOS representation, phrase structure is defined over eojeol terminals, but each eojeol retains fine-grained Korean POS information through a combined XPOS preterminal. In the Eojeol+UPOS representation, the same eojeol terminal domain is paired with a coarser universal POS layer. This design allows us to ask whether parsing differences come from the terminal domain itself or from the amount of morphosyntactic information available at the preterminal level.

The scope of this paper is deliberately limited to the Penn Korean Treebank. Other Korean constituency resources raise different conversion problems. The Sejong treebank would require principled debinarization before direct comparison with non-binary phrase structures. The KAIST treebank has been adapted for surface-oriented parsing in the SPMRL shared-task resources \citep{seddah-EtAl:2013:SPMRL,seddah-kubler-tsarfaty:2014}, but starting from the original KAIST representation would require a separate theoretical and conversion study. The issue is not merely grouping morphemes into eojeol terminals. KAIST-style analysis reflects theory-driven phrase structure influenced by Government and Binding approaches \citep{chomsky:1981,chomsky-1986-barriers}, where functional morphemes may be represented as syntactically relevant units. An eojeol-based conversion would therefore have to decide whether such functional structure should be preserved in phrase structure, flattened into eojeol terminals, or reassigned to a separate morphological layer. These are linguistic decisions, not only preprocessing choices. We therefore use the Penn Korean Treebank as a conservative source for comparing representations derived from the same phrase-structural analysis.

On these representations, we evaluate three canonical non-binary transition-based constituency parsers: top-down \citep{dyer-etal-2016-recurrent}, in-order \citep{liu-zhang-2017-order}, and bottom-up \citep{gonzalez-rodriguez-2019-faster}. The systems construct the same class of non-binary trees but differ in when constituent labels are introduced. Top-down parsing predicts a phrase label before its children are built; in-order parsing introduces the label after the left corner; and bottom-up parsing delays label prediction until the child sequence has been constructed. These differences are relevant for Korean because important morphosyntactic cues often occur at the right edge of an eojeol or {a} clause. By implementing the three systems under the same architecture, we examine how transition order interacts with terminal granularity and morphosyntactic encoding.

Previous work on Korean constituency parsing has used different treebanks, tokenization assumptions, and parsing models. Early work includes statistical parsing of a Korean newspaper treebank \citep{hermjakob:2000:NAACL,hermjakob-mooney:1997:EACL}, LTAG parsing from the early Penn Korean Treebank \citep{sarkar-han:2002:TAG,sarkar:2002,xia-EtAl:2000:CLPW,bangalore-joshi:1999:CL}, and PCFG-based parsing of the Penn Korean Treebank \citep{chung-post-gildea:2010:SPMRL}. Constituency parsing has also been studied with the Sejong treebank \citep{park:2006,choi-park-choi:2012:SP-SEM-MRL,park-hong-cha:2016:PACLIC,kim-park:2022} and with the modified KAIST treebank used in the SPMRL shared tasks \citep{seddah-EtAl:2013:SPMRL,seddah-kubler-tsarfaty:2014,bjorkelund-EtAl:2013:SPMRL,bjorkelund-EtAl:2014:SPMRL}. More recent neural constituency parsers provide strong general architectures \citep{vinyals-etal-2015-grammar,kuncoro-etal-2017-recurrent,kitaev-klein-2018-constituency,kitaev-cao-klein:2019:ACL,fried-nikita-klein:2019:ACL}, but the representational question of Korean terminal granularity remains independent of the parsing model itself.

The contributions of this paper are threefold. First, we define a conservative conversion procedure that derives aligned morpheme-based and eojeol-based constituency representations from the Penn Korean Treebank while preserving the source phrase-structural analysis where possible. Second, we compare Morpheme+XPOS, Eojeol+XPOS, and Eojeol+UPOS parsing targets to examine how terminal and POS granularity affect Korean constituency parsing under gold segmentation and preterminal annotation. Third, we evaluate top-down, in-order, and bottom-up non-binary transition systems across these representations, analyzing overall accuracy, transition length, and arity-specific constituent recovery.

The empirical and representational claims should be distinguished. The experiments show that richer morphological representations provide valuable evidence for the evaluated transition-based parsers and improve recovery of eojeol-level phrase structure under gold annotation. This finding concerns the information available to the parser; it does not independently establish the linguistically preferable terminal domain.

The proposal to use eojeol as the primary constituency terminal domain is instead based on the linguistic and resource-design considerations developed later in this paper and in related work \citep{park-park-2026-constituency}. Under this proposal, eojeols provide an overt and interpretable surface domain for constituency structure, while morpheme sequences and XPOS labels remain available as aligned morphosyntactic annotation. An eojeol-primary design also facilitates alignment with other Korean language resources, including Universal Dependencies \citep{chun-EtAl:2018:LREC,noh-etal-2018-enhancing,seo-etal-2019-ud,kim-etal-2024-kud}, PropBank,\footnote{\url{https://doi.org/10.35111/j0yk-ph77}} and FrameNet \citep{park-EtAl:2014:ISWCPOSTER,hahm-EtAl:2018:LREC}, where annotation is typically anchored to eojeol-level units while finer-grained morphosyntactic information can be maintained in an associated layer.

\section{Background}

\subsection{Korean constituency parsing and terminal granularity}

The choice of terminal units is central to Korean constituency parsing. In English-style constituency parsing, terminals typically correspond to surface word tokens, and preterminals provide part-of-speech labels immediately above those tokens. Korean presents a more complex case because {eojeol, its orthographic spacing unit,} often contains several morphemes. A single eojeol may combine a lexical stem with case particles, auxiliary material, verbal endings, derivational suffixes, or punctuation. As a result, constituency parsing for Korean must decide whether phrase structure is built over eojeol-level surface units or over morpheme-level units obtained through morphological analysis.

This issue is part of a broader problem of word segmentation granularity in Korean language processing. Korean NLP resources and systems do not always agree on whether the basic processing unit should be the eojeol, the morpheme, or some intermediate unit. Previous work on Korean segmentation granularity has shown that different segmentation choices affect how lexical, morphological, and syntactic information is represented in downstream tasks \citep{park-kim-2024-word}. Constituency parsing makes this issue especially explicit because the segmentation choice determines the terminal yield of the tree itself. Terminal granularity is therefore not only an input preprocessing decision; it changes the formal object that the parser is trained to produce.

This decision affects both the linguistic interpretation of trees and the practical design of parsers. If morphemes are treated as constituency terminals, phrase structure is defined over a morphologically explicit sequence. This representation exposes case particles, verbal endings, derivational suffixes, and other functional morphemes as terminal-level material, but it also makes the parsing target dependent on morphological segmentation and fine-grained Korean POS annotation. Conversely, if {eojeols} are treated as terminals, phrase structure is defined over the surface spacing units available in ordinary Korean text, while internal morphological information must be represented elsewhere, for example through preterminal labels, morphological features, or an aligned auxiliary layer.

The distinction between eojeol-based and morpheme-based representations should not be understood as a simple choice between linguistically informed and linguistically uninformed parsing. Both representations encode linguistic information, but they place it in different parts of the treebank. A morpheme-based representation projects morphology into the terminal yield. An eojeol-based representation keeps the terminal yield closer to the overt sentence and must encode morphology through labels or aligned annotation. The present paper therefore treats terminal granularity and POS granularity as related but distinct representational choices.

The eojeol-based view is useful for {aligning} constituency parsing with existing Korean dependency resources, since Korean dependency treebanks are commonly organized over eojeol-level tokens. An eojeol-based constituency treebank provides a shared terminal domain for constituency–dependency comparison and conversion. At the same time, a morpheme-based representation may provide richer local morphosyntactic cues to the parser because case particles, endings, and derivational elements are available as independent terminals. This tension motivates the comparison in this paper: we ask how parsing behavior changes when the same source phrase-structural analysis is represented with Morpheme+XPOS, Eojeol+XPOS, and Eojeol+UPOS targets.

\subsection{The Penn Korean Treebank}

The Penn Korean Treebank \citep{han-EtAl:2002} is one of the major constituency resources for Korean. It adapts the Penn Treebank annotation tradition to Korean, using phrase labels and functional annotations to represent nominal, verbal, clausal, and adjunct structures. Like other Penn-style resources, it also includes explicit null elements, which are used to represent unrealized arguments and other empty categories.

The treebank is particularly valuable because it provides rich phrase-structural analyses of Korean sentences. It includes labels such as \texttt{NP}, \texttt{VP}, and \texttt{S}, together with functional extensions such as subject, object, adjunct, and complement marking. These annotations make it possible to study Korean phrase structure in a way that is more explicit than dependency annotation alone.

At the same time, the original treebank is not immediately identical to any of the three parsing targets compared in this paper. Its terminals contain morphological analyses rather than plain surface eojeol, and null elements appear as structural leaves even though they do not correspond to overt input tokens. In addition, some functional morphology is represented in canonicalized form, abstracting away from surface allomorphy. These properties are linguistically meaningful within the original annotation scheme, but they create a representational gap between the original treebank and surface-oriented parsing targets.

For this reason, we use the Penn Korean Treebank as a controlled source from which several related parsing representations can be derived. The choice is deliberate. Other Korean constituency resources, such as Sejong and KAIST, raise additional conversion questions before they can support the same kind of comparison. Sejong requires principled debinarization of its distributed phrase-structure annotation. {KAIST raises theoretically motivated questions about how its original structural analysis should be interpreted when the terminal domain is changed}. These issues are important, but they would introduce further variables beyond the representation comparison pursued here.

By contrast, the Penn Korean Treebank provides a comparatively conservative starting point. Its overt terminals are already associated with eojeol-level units with terminal-internal morphological annotation. After null-element removal, the remaining overt yield can be aligned with an eojeol sequence. Each eojeol can then be associated with its surface form, morpheme sequence, XPOS annotation, and UPOS label. This aligned representation allows us to extract the three parsing targets used in the experiments: Morpheme+XPOS, Eojeol+XPOS, and Eojeol+UPOS.


Sentence~(\ref{penn-1}) illustrates the kind of representation provided by the Penn Korean Treebank. The sentence is annotated with phrase structure over eojeol-level units that contain terminal-internal morphological analysis. The example also contains a null subject \texttt{pro} {inside the complement clause}. This combination is characteristic of the resource: overt terminals are already grouped as eojeol-like units, but each terminal records its internal morpheme sequence and fine-grained Korean POS tags, and non-overt elements may appear as ordinary leaves in the tree.


\begin{exe}
\ex \label{penn-1}
\begin{xlist}
\ex \label{penn-sentence-1}
\glll 영국정부는 2 일 중 전 칠레 독재자 아우구스토 피노체트의 최종 처리 방침을 발표할 것이라고 밝혔다. \\
\textit{yeongguk}\textit{jeongbuneun} 2 \textit{il} \textit{jung} \textit{jeon} \textit{chille} \textit{dokjaeja} \textit{auguseuto} \textit{pinocheteui} \textit{choejong} \textit{cheori} \textit{bangchimeul} \textit{balpyohal} \textit{geosirago} \textit{balkyeotda.} \\
UK.government-\textsc{top} two day during former Chile dictator Augusto Pinochet-\textsc{gen} final handling policy-\textsc{acc} announce-\textsc{fut} thing-\textsc{comp} state-\textsc{pst}.\textsc{decl} \\
\trans `The British government stated that on the 2nd it would announce its final policy on handling former Chilean dictator Augusto Pinochet.’

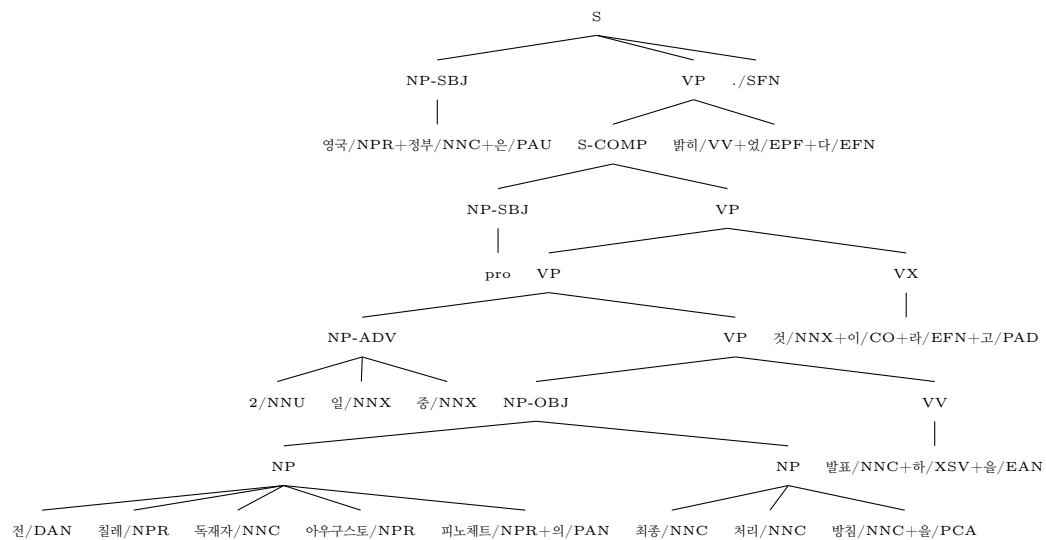
\ex \label{penn-original-tree-1}
\tiny{
{
\begin{forest}
[S [NP-SBJ [영국/NPR+정부/NNC+은/PAU]]
[VP [S-COMP [NP-SBJ [pro]] [VP [VP [NP-ADV [2/NNU] [일/NNX] [중/NNX]]
[VP [NP-OBJ [NP [전/DAN] [칠레/NPR] [독재자/NNC] [아우구스토/NPR] [피노체트/NPR+의/PAN]]
[NP [최종/NNC] [처리/NNC] [방침/NNC+을/PCA]]]
[VV [발표/NNC+하/XSV+을/EAN]]]]
[VX [것/NNX+이/CO+라/EFN+고/PAD]]]]
[밝히/VV+었/EPF+다/EFN]] [${.}$/SFN]]
\end{forest}}
}
\end{xlist}
\end{exe}

\section{Morpheme- and eojeol-based treebank representations}
\label{sec:representations}

This section describes the three constituency treebank representations compared in this paper. The goal is not to propose a single replacement for the original Penn Korean Treebank, but to make explicit how different choices about terminal granularity and POS granularity change the parsing target. We therefore distinguish between morpheme-based and eojeol-based constituency representations, and within the eojeol-based setting, between fine-grained XPOS and coarse-grained UPOS preterminal annotation.

The three representations are derived from the same Penn Korean phrase-structural source. They therefore share the same sentence set, the same underlying phrase labels where possible, and the same source analysis of Korean phrase structure. They differ in how the terminal yield is defined and how morphosyntactic information associated with each Korean spacing unit is represented. The Morpheme+XPOS representation treats internal morphemes as constituency terminals. The Eojeol+XPOS and Eojeol+UPOS representations treat overt {eojeols} as constituency terminals, but differ in the category assigned above each eojeol.

\subsection{Aligned phrase-structural source}

The Penn Korean Treebank provides phrase-structural brackets and Korean-specific phrase labels, while the aligned CoNLL-U representation provides overt eojeol tokens, morpheme sequences, XPOS tags, and UPOS labels. We use this alignment to view each overt eojeol as a unit with two kinds of information: a position in the Penn Korean phrase-structural tree and an internal morphological analysis.

Each morpheme is assigned through this gold alignment to exactly one eojeol. This fixed morpheme-to-eojeol mapping is retained across the three representations and is used to project Morpheme+XPOS predictions into the eojeol span domain for normalized evaluation (Section~\ref{sec:normalization}).

For example, an eojeol such as 이르면 \textit{ireumyeon} (`if it is early’) may be associated with the morpheme sequence \texttt{이르+으면}, the XPOS sequence \texttt{VJ+ECS}, and the UPOS label \texttt{ADJ}. These annotations provide alternative ways of representing the same surface unit in a constituency tree. In a morpheme-based representation, the internal morphemes are projected as terminals. In an eojeol-based representation, the eojeol remains a single terminal, while the associated POS information is encoded in the preterminal layer.


The alignment is therefore not merely a preprocessing device. It defines a common representational basis from which multiple parsing targets can be extracted. This makes it possible to compare treebank representations while holding the source phrase-structural analysis constant.

\subsection{Morpheme-based representation}

In the morpheme-based representation, constituency terminals correspond to morphemes, and preterminals correspond to fine-grained Korean XPOS tags. An eojeol is expanded into its internal morpheme sequence, and each morpheme is represented as a terminal dominated by its XPOS preterminal. Thus, an eojeol such as 영국정부는 \textit{yeonggukjeongbuneun} (`the British government-\textsc{top}’) may be represented as the sequence 영국 \textit{yeongguk} with \texttt{NPR}, 정부 \textit{jeongbu} with \texttt{NNC}, and 은 \textit{eun} with \texttt{PAU}.

This representation makes Korean morphology directly visible in the terminal yield of the constituency tree. Case particles, verbal endings, derivational suffixes, auxiliary material, and other functional morphemes can appear as independent terminal-level items. As a result, syntactically relevant morphological cues are available to the parser as part of the tree structure itself. For example, particles marking topic, nominative, accusative, or genitive function are represented as separate terminals, and endings involved in clausal subordination, adnominal modification, or complementation can be distinguished by their XPOS labels.

The morpheme-based representation therefore reflects a morphologically explicit view of Korean constituency structure. Its terminal sequence is longer than the corresponding eojeol sequence, and the parser must build phrase structure over a more fine-grained yield. At the same time, the representation gives the parser direct access to fine-grained morphosyntactic information that may be crucial for identifying phrase labels and constituent boundaries in Korean.

\subsection{Eojeol-based representation}

In the eojeol-based representation, constituency terminals correspond to overt {eojeols}. Internal morphemes are not projected as independent terminals of the constituency tree. Instead, the eojeol is treated as the surface unit over which phrase-structural spans are defined. For example, 영국정부는 \textit{yeonggukjeongbuneun} is represented as a single terminal rather than as three separate morpheme terminals.

{This representation separates two questions that are often conflated: what counts as the terminal unit of the constituency tree, and where morphosyntactic information is represented. The goal is not to exclude morphology, but to avoid making morphemes themselves the obligatory terminal units of constituency structure. Morpheme-level information can instead be retained in an aligned layer or encoded through preterminal features.}

The eojeol-based representation has a shorter terminal sequence and defines constituent spans over the overt spacing units of the sentence. This makes it compatible with the surface tokenization used in many Korean dependency resources and downstream applications. It also makes the phrase-structural yield correspond directly to the spacing units encountered in ordinary Korean text. However, this surface orientation raises a representational question: how much of the internal morphosyntactic information of the eojeol should be encoded in the constituency tree itself?

We compare two answers to this question. In the Eojeol+XPOS representation, each eojeol is a single terminal, but its preterminal label records the combined fine-grained XPOS sequence of the morphemes inside the eojeol. Thus, an eojeol such as 이르면 \textit{ireumyeon} may be represented as a single terminal dominated by a preterminal such as \texttt{VJ+ECS}. This preserves detailed Korean morphosyntactic information while keeping the terminal domain eojeol-based.

In the Eojeol+UPOS representation, each eojeol is again a single terminal, but its preterminal label is a coarse UPOS category such as \texttt{NOUN}, \texttt{PROPN}, \texttt{VERB}, \texttt{ADJ}, \texttt{ADV}, or \texttt{PUNCT}. This representation abstracts away from the internal XPOS sequence and assigns a single syntactic category to the eojeol as a whole. It therefore provides a smaller and more cross-linguistically comparable preterminal inventory, but it also removes many distinctions expressed by Korean-specific morphology.

The difference between Eojeol+XPOS and Eojeol+UPOS separates terminal granularity from POS granularity. Both representations define phrase structure over the same eojeol terminals, but they differ in how much morphosyntactic information is visible at the preterminal level. Comparing these two settings allows us to ask whether difficulties in eojeol-based parsing arise from the eojeol terminal domain itself, from the loss of fine-grained POS information, or from the interaction between the two.

\subsection{Three parsing targets}

The three representations compared in this paper can be summarized as follows. In the Morpheme+XPOS representation, terminals are morphemes and preterminals are fine-grained Korean POS tags. In the Eojeol+XPOS representation, terminals are {eojeols} and preterminals are combined XPOS sequences. In the Eojeol+UPOS representation, terminals are {eojeols} and preterminals are coarse UPOS labels.

\begin{table}
\centering
\footnotesize
\begin{tabular}{llll}
\toprule
Representation & Terminal unit & Preterminal label & Morphological information \\
\midrule
Morpheme+XPOS & morpheme & XPOS & Projected into terminal yield \\
Eojeol+XPOS & Eojeol & Combined XPOS sequence & Encoded in preterminal label \\
Eojeol+UPOS & Eojeol & UPOS & Abstracted to coarse category \\
\bottomrule
\end{tabular}
\caption{Three constituency treebank representations compared in this paper.}
\label{tab:three-representations}
\end{table}

These representations differ in where they place morphosyntactic information. The Morpheme+XPOS representation places it in the terminal sequence. The Eojeol+XPOS representation keeps the terminal sequence surface-oriented but preserves detailed morphology in the preterminal. The Eojeol+UPOS representation keeps both the terminal and preterminal layers more compact, but represents only a coarse category for each eojeol. The comparison therefore evaluates not only morpheme-based versus eojeol-based parsing, but also the role of POS granularity within eojeol-based parsing.

\subsection{Illustration}

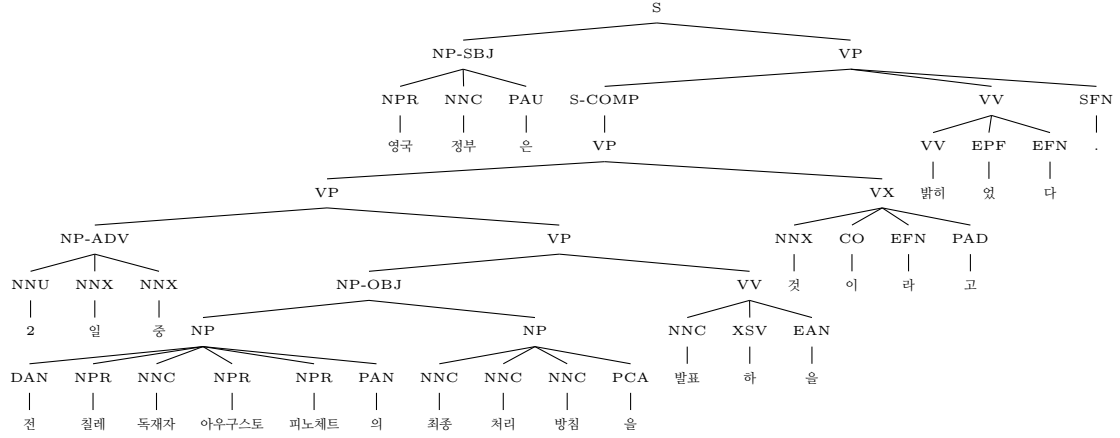
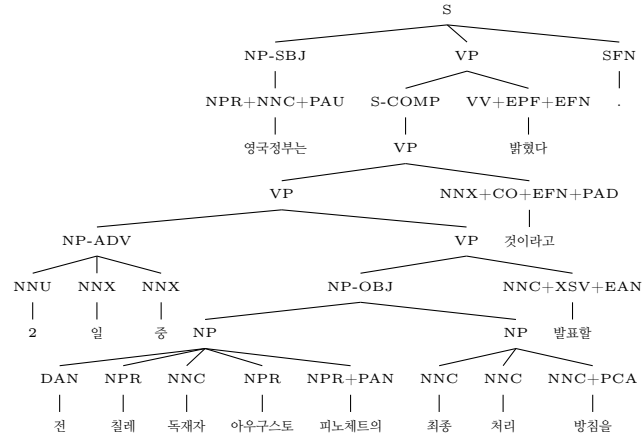
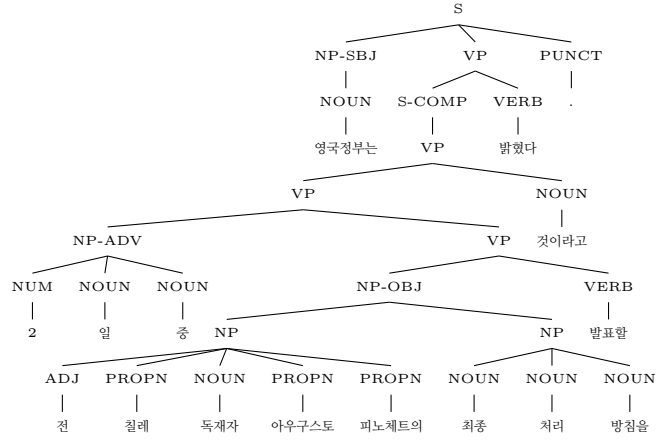
\begin{figure}
\centering

\begin{subfigure}[t]{\textwidth}
\centering
{
\tiny{\begin{forest}
[S
[NP-SBJ
[NPR [영국]]
[NNC [정부]]
[PAU [은]]]
[VP
[S-COMP
[VP
[VP
[NP-ADV
[NNU [2]]
[NNX [일]]
[NNX [중]]]
[VP
[NP-OBJ
[NP
[DAN [전]]
[NPR [칠레]]
[NNC [독재자]]
[NPR [아우구스토]]
[NPR [피노체트]]
[PAN [의]]]
[NP
[NNC [최종]]
[NNC [처리]]
[NNC [방침]]
[PCA [을]]]]
[VV
[NNC [발표]]
[XSV [하]]
[EAN [을]]]]]
[VX
[NNX [것]]
[CO [이]]
[EFN [라]]
[PAD [고]]]]]
[VV
[VV [밝히]]
[EPF [었]]
[EFN [다]]]
[SFN [${.}$]]]]
\end{forest}}
}
\caption{morpheme terminals with XPOS preterminals.}\label{fig:penn-morpheme-xpos-tree}
\end{subfigure}

\begin{subfigure}[t]{\textwidth}
\centering
{
\tiny{\begin{forest}
[S
[NP-SBJ
[NPR+NNC+PAU [영국정부는]]]
[VP
[S-COMP
[VP
[VP
[NP-ADV
[NNU [2]]
[NNX [일]]
[NNX [중]]]
[VP
[NP-OBJ
[NP
[DAN [전]]
[NPR [칠레]]
[NNC [독재자]]
[NPR [아우구스토]]
[NPR+PAN [피노체트의]]]
[NP
[NNC [최종]]
[NNC [처리]]
[NNC+PCA [방침을]]]]
[NNC+XSV+EAN [발표할]]]]
[NNX+CO+EFN+PAD [것이라고]]]]
[VV+EPF+EFN [밝혔다]]]
[SFN [${.}$]]]
\end{forest}}
}
\caption{Eojeol terminals with combined XPOS preterminals.}\label{fig:penn-eojeol-xpos-tree}
\end{subfigure}


\begin{subfigure}[t]{\textwidth}
\centering
{
\tiny{\begin{forest}
[S
[NP-SBJ
[NOUN [영국정부는]]]
[VP
[S-COMP
[VP
[VP
[NP-ADV
[NUM [2]]
[NOUN [일]]
[NOUN [중]]]
[VP
[NP-OBJ
[NP
[ADJ [전]]
[PROPN [칠레]]
[NOUN [독재자]]
[PROPN [아우구스토]]
[PROPN [피노체트의]]]
[NP
[NOUN [최종]]
[NOUN [처리]]
[NOUN [방침을]]]]
[VERB [발표할]]]]
[NOUN [것이라고]]]]
[VERB [밝혔다]]]
[PUNCT [${.}$]]]
\end{forest}}
}
\caption{Eojeol terminals with UPOS preterminals.}\label{fig:penn-eojeol-upos-tree}
\end{subfigure}

\caption{Three terminal/POS representations derived from the same sentence.}\label{fig:three-representation-trees}
\end{figure}

Sentence~(\ref{penn-1}) illustrates the Penn Korean source representation used to derive the three parsing targets. The original tree contains eojeol-like terminals with internal morphological annotation, such as \texttt{영국/NPR+정부/NNC+은/PAU}, and it also contains non-overt material such as the null subject \texttt{pro}. 
Figure~\ref{fig:three-representation-trees} illustrates the three input representations: morpheme-based, Eojeol+XPOS, and Eojeol+UPOS. Section~\ref{sec:experiments} compares them under the same parsing architecture and evaluation protocol.

\section{Canonical non-binary transition systems}
\label{sec:canonical-transition-systems}

This section defines the three canonical non-binary transition systems used in
the experiments. All three systems construct projective constituency trees over
the input sequence, but they differ in the order in which constituent labels
and child sequences are introduced. Top-down parsing introduces a phrase label
before its children are constructed. In-order parsing introduces the label
after the left corner has been built. Bottom-up parsing constructs the child
sequence first and introduces the mother label at reduction time.

A parser configuration is written as
\[
  c = \langle \sigma, i, f \rangle ,
\]
where \(\sigma\) is a stack, \(i\) is the index of the next input token, and
\(f \in \{0,1\}\) is a completion flag. For an input sentence
\(w=w_1\cdots w_n\), the initial configuration is
\[
  \langle \epsilon,1,0\rangle ,
\]
and a successful final configuration has the form
\[
  \langle T,n+1,1\rangle ,
\]
where \(T\) is a complete constituency tree over \(w\). Stack concatenation is
written with \(\mid\), so \(\sigma\mid x\) denotes the stack obtained by
pushing \(x\) onto \(\sigma\). 
We write completed tree items as
\[
X(\alpha_1,\ldots,\alpha_k),
\]
where \(X\) is a constituent label and
\(\alpha_1,\ldots,\alpha_k\) are its ordered children.

The converted eojeol-based treebank is not binarized. Many constituents have
more than two children, especially in nominal sequences, modifier structures,
and clausal projections. Non-binary transition systems are therefore a natural
choice because they can build such constituents directly, without introducing
artificial binary intermediate nodes.

\subsection{Top-down non-binary parsing}
\label{sec:background-topdown}

In top-down non-binary parsing, a constituent label is introduced before any of
its children are constructed. The canonical example is the transition system
used in recurrent neural network grammars \citep{dyer-etal-2016-recurrent}. A
nonterminal marker \(\textsc{nt}(X)\) records that a constituent labeled \(X\)
has been opened before its children are built.

The transition inventory is:
\[
\begin{array}{llll}
\textsc{Shift}: &
  \langle \sigma,\ i,\ f \rangle
  & \Rightarrow &
  \langle \sigma \mid w_i,\ i+1,\ f \rangle ,
  \quad i\leq n,
\\[1ex]

\textsc{Nt}(X): &
  \langle \sigma,\ i,\ f \rangle
  & \Rightarrow &
  \langle \sigma \mid \textsc{nt}(X),\ i,\ f \rangle ,
\\[1ex]

\textsc{Reduce}: &
  \langle \sigma \mid \textsc{nt}(X) \mid
          \alpha_1 \mid \cdots \mid \alpha_k,\ i,\ f \rangle
  & \Rightarrow &
  \langle \sigma \mid X(\alpha_1,\ldots,\alpha_k),\ i,\ f \rangle ,
  \quad k\geq 1,
\\[1ex]

\textsc{Finish}: &
  \langle T,\ n+1,\ 0 \rangle
  & \Rightarrow &
  \langle T,\ n+1,\ 1 \rangle .
\end{array}
\]

Here \(\alpha_1,\ldots,\alpha_k\) are the ordered children of the constituent.
The \(\textsc{Nt}(X)\) action places a nonterminal marker on the stack before
any of these children are constructed. The \(\textsc{Reduce}\) action then
combines all items above the most recent unmatched marker into a constituent
labeled \(X\). When \(k=1\), the reduction produces a unary constituent;
\(k=2\) produces a binary constituent; and \(k\geq3\) produces a higher-arity
constituent. The reduction action is not arity-specific, since its arity is
determined by the number of items above the marker.

\subsection{In-order non-binary parsing}
\label{sec:background-inorder}

In-order non-binary parsing introduces a constituent label after its
left-corner child has been constructed but before its remaining children are
constructed \citep{liu-zhang-2017-order}. A nonterminal marker
\(\textsc{nt}(X)\) records that a constituent labeled \(X\) has been opened
after recognition of its left corner.

The transition inventory is:
\[
\begin{array}{llll}
\textsc{Shift}: &
  \langle \sigma,\ i,\ f \rangle
  & \Rightarrow &
  \langle \sigma \mid w_i,\ i+1,\ f \rangle ,
  \quad i\leq n,
\\[1ex]

\textsc{Nt}(X): &
  \langle \sigma \mid \alpha_1,\ i,\ f \rangle
  & \Rightarrow &
  \langle \sigma \mid \alpha_1 \mid \textsc{nt}(X),\ i,\ f \rangle ,
\\[1ex]

\textsc{Reduce}: &
  \langle \sigma \mid \alpha_1 \mid \textsc{nt}(X) \mid
          \alpha_2 \mid \cdots \mid \alpha_k,\ i,\ f \rangle
  & \Rightarrow &
  \langle \sigma \mid X(\alpha_1,\alpha_2,\ldots,\alpha_k),\ i,\ f \rangle ,
  \quad k\geq 1,
\\[1ex]

\textsc{Finish}: &
  \langle T,\ n+1,\ 0 \rangle
  & \Rightarrow &
  \langle T,\ n+1,\ 1 \rangle .
\end{array}
\]

Here \(\alpha_1\) is the left-corner child, and
\(\alpha_2,\ldots,\alpha_k\) are the remaining ordered children of the
constituent. The \(\textsc{Nt}(X)\) action places a nonterminal marker
immediately above the left corner, before the remaining children are
constructed. The \(\textsc{Reduce}\) action then combines the left corner and
all items above the marker into a constituent labeled \(X\). When \(k=1\), the
reduction produces a unary constituent; \(k=2\) produces a binary constituent;
and \(k\geq3\) produces a higher-arity constituent. As in the top-down system,
the reduction action is not arity-specific, since its arity is determined by
the stack configuration.

\subsection{Bottom-up non-binary parsing with explicit arity}
\label{sec:background-bottomup}

Bottom-up non-binary parsing introduces a constituent label only after all of
its children have been constructed. In the canonical arity-specific
formulation, no open nonterminal marker identifies the left boundary of the
future constituent. The label and reduction scope are therefore encoded
directly in the reduction action.

We use the arity-specific non-binary bottom-up system of
\citet{gonzalez-rodriguez-2019-faster}. Its transition inventory is:
\[
\begin{array}{llll}
\textsc{Shift}: &
  \langle \sigma,\ i,\ f \rangle
  & \Rightarrow &
  \langle \sigma \mid w_i,\ i+1,\ f \rangle ,
  \quad i\leq n,
\\[1ex]

\textsc{Reduce}\text{-}X\#k: &
  \langle \sigma \mid \alpha_1 \mid \cdots \mid \alpha_k,\ i,\ f \rangle
  & \Rightarrow &
  \langle \sigma \mid X(\alpha_1,\ldots,\alpha_k),\ i,\ f \rangle ,
  \quad k\geq 1,
\\[1ex]

\textsc{Finish}: &
  \langle T,\ n+1,\ 0 \rangle
  & \Rightarrow &
  \langle T,\ n+1,\ 1 \rangle .
\end{array}
\]

Here \(\alpha_1,\ldots,\alpha_k\) are the ordered children of the constituent.
The \(\textsc{Reduce}\text{-}X\#k\) action pops exactly the top \(k\) stack
items, combines them into a constituent labeled \(X\), and pushes the completed
constituent onto the stack. When \(k=1\), the action produces a unary
constituent; \(k=2\) produces a binary constituent; and \(k\geq3\) produces a
higher-arity constituent. Unlike the top-down and in-order systems, the
bottom-up system has no nonterminal marker that determines the reduction
domain. Its reduction actions are therefore arity-specific, with a distinct
action for each observed combination of constituent label \(X\) and arity
\(k\).

\section{Experiments}
\label{sec:experiments}

This section describes the experimental setup and reports the main parsing results. The goal is to compare top-down, in-order, and bottom-up non-binary transition systems across three aligned representations derived from the same Penn Korean Treebank sentences: Morpheme+XPOS, Eojeol+XPOS, and Eojeol+UPOS. All systems use the same data splits, neural architecture, training procedure, and evaluation protocol, while receiving the gold terminal sequence and preterminal labels defined by their respective representations.

\subsection{Data splits}
\label{sec:data-splits}

We use the Penn Korean Treebank. The original trees are converted by removing
null elements following \citet{kitaev-klein-2018-constituency},\footnote{\url{https://github.com/nikitakit/self-attentive-parser/tree/master/data/common}}
aligning Penn Korean phrase structure with overt eojeol tokens using its UD
counterpart,\footnote{\url{https://catalog.ldc.upenn.edu/LDC2023T05}}
and preserving Penn Korean phrase labels where possible. From this conversion,
we derive three parsing representations: Morpheme+XPOS, Eojeol+XPOS, and
Eojeol+UPOS. In the Morpheme+XPOS representation, morphemes are terminals and
XPOS tags are preterminals. In the Eojeol+XPOS representation, {eojeols} are
terminals and combined XPOS sequences are preterminals. In the Eojeol+UPOS
representation, {eojeols} are terminals and UPOS tags are preterminals. Examples
of the three representations are shown in
Figures~\ref{fig:penn-morpheme-xpos-tree}--\ref{fig:penn-eojeol-upos-tree}.

We divide the converted treebank into training, development, and test sets
using a deterministic file-level split.\footnote{The Penn Korean Treebank has
rarely been used for constituency parsing. One exception is
\citet{chung-post-gildea:2010:SPMRL}, who used all 5,010 sentences of the
treebank and evaluated with ten-fold cross validation. In contrast, we use a
fixed file-level split in order to keep all sentences from the same source file
in the same partition.}
The split is based on the final digit of the Penn Korean Treebank file
identifier. Files whose identifiers end in 1--8 are assigned to the training
set, files ending in 9 are assigned to the development set, and files ending in
0 are assigned to the test set. This gives an approximately 80/10/10 split at
the file level while avoiding sentence-level mixing across partitions.

Table~\ref{tab:data-splits} summarizes the resulting data splits for the three
representations. For each split, we report the number of files, sentences, and
terminals, where terminals correspond to morphemes in the Morpheme+XPOS
representation and {eojeols} in the Eojeol+XPOS and Eojeol+UPOS representations.

\begin{table}
\centering
\footnotesize
\begin{tabular}{lccc}
\toprule
Split
& Files
& Sentences
& Terminals \\
\midrule
Train & 85 & 3,877 & 189,117 / 102,613 \\
Dev   & 9  & 428   & 21,133 / 11,435 \\
Test  & 18 & 685   & 31,631 / 17,354 \\
\bottomrule
\end{tabular}
\caption{Statistics of the converted Penn Korean Treebank under the fixed
file-level split. Files and sentences are shared across the three
representations. The terminal counts are reported as morphemes/{eojeols}: the
first value corresponds to the Morpheme+XPOS representation, and the second to
the Eojeol+XPOS and Eojeol+UPOS representations. Files ending in 1--8 are used
for training, files ending in 9 for development, and files ending in 0 for
testing.}
\label{tab:data-splits}
\end{table}

\subsection{Parser implementation details}

All experiments use gold terminal segmentation and gold preterminal labels as parser input. Accordingly, the comparison evaluates constituency parsing conditioned on gold tokenization and preterminal annotation rather than end-to-end parsing from surface text alone. In Morpheme+XPOS, the encoder receives the gold morpheme sequence and the gold XPOS label of each morpheme. In Eojeol+XPOS, it receives the gold eojeol sequence and the gold combined XPOS sequence assigned to each eojeol. In Eojeol+UPOS, it receives the gold eojeol sequence and the gold UPOS label assigned to each eojeol. Preterminals are therefore supplied rather than predicted, accounting for the essentially 100\% scores at tree height 1.

All systems are implemented within the transition-based constituency parsing framework provided by Stanza \citep{bauer-manning-2025-high}.\footnote{\url{https://github.com/stanfordnlp/stanza/tree/main/stanza/models/constituency}} Although Stanza includes top-down and in-order transition systems, we independently implement all three systems compared here—top-down, in-order, and arity-specific bottom-up—using the same Stanza transition interfaces and neural parsing framework. For each system, we implement its static oracle, transition inventory, applicability constraints, and transition-application procedure.

The shared model is Stanza's recurrent constituency parser. The input sentence is encoded by a bidirectional LSTM. Each input item is represented by the concatenation of a trainable word embedding, a trainable preterminal-label embedding, a fixed pretrained word vector, and representations from fixed forward and backward character language models. We use the Korean pretrained word vectors and character language models distributed with Stanza, without contextualized transformer representations. The transition history and constructed constituents are represented by LSTM stacks. We otherwise retain the standard Stanza constituency-parser configuration and use the same hyperparameter settings in every experimental condition.

Except for the transition-specific components, the sentence encoder, parser-state representation, subtree-composition function, transition scorer, optimization procedure, checkpoint-selection criterion, and decoding settings are identical across all systems and representations. Each parser is trained with its transition-specific static oracle, and greedy decoding selects the highest-scoring structurally applicable transition.

\subsection{Normalization to Eojeol+UPOS}
\label{sec:normalization}

To compare the three representations in a common span domain, we convert Morpheme+XPOS and Eojeol+XPOS predictions to Eojeol+UPOS trees before normalized evaluation. For Morpheme+XPOS, the conversion uses the aligned CoNLL-U representation as a pivot. The CoNLL-U FORM and UPOS fields supply the eojeol terminal and preterminal label, while its morphological analysis determines which consecutive morphemes belong to each eojeol. The conversion therefore uses gold eojeol boundaries and gold morpheme-to-eojeol alignments rather than inferring them from the predicted phrase structure.

The converter first verifies that the predicted tree has the same morpheme yield as the aligned CoNLL-U sentence and then computes the morpheme span of each predicted constituent. A constituent is retained if both endpoints coincide with eojeol boundaries. A node whose left or right endpoint falls within an eojeol is removed, and its surviving children are promoted to its parent. Constituents spanning multiple eojeols are therefore preserved when both endpoints align with eojeol boundaries, whereas structure that cannot be expressed over an eojeol terminal sequence is discarded.

The morpheme terminals and XPOS preterminals within each eojeol are then replaced by a single eojeol terminal with its gold UPOS preterminal. Boundary-aligned unary chains are retained; if multiple nodes project to the same span, their original dominance order is preserved. The conversion does not silently repair malformed predictions: a tree must match the aligned morpheme yield and produce every eojeol exactly once in its original order.

For Eojeol+XPOS predictions, no span projection or terminal regrouping is required. The eojeol terminals and predicted phrase structure are preserved, while the combined gold XPOS preterminals are replaced by the corresponding gold UPOS preterminals. Native Eojeol+UPOS predictions require no conversion. All three conditions are then evaluated against the same Eojeol+UPOS gold representation.

Projection can slightly increase labeled \(F_1\), as in Table~\ref{tab:converted-results}, because the native and normalized evaluations score different representational domains. Morpheme-internal nodes and constituents with boundaries inside eojeols are excluded after projection, so errors confined to distinctions that cannot be expressed over eojeol spans no longer enter the normalized evaluation. The converted score therefore measures recovery of eojeol-level constituency structure rather than improvement to the original morpheme-level prediction.

\subsection{Evaluation metrics}
\label{sec:evaluation-metrics}

We report corpus-level labeled bracketing \(F_1\) using both \texttt{EVALB} \citep{black-etal-1991-procedure} and \texttt{jp-evalb} \citep{jo-etal-2024-novel}. The scores are averaged over constituents in the test set. Standard \texttt{EVALB} applies conventional normalization rules, including punctuation treatment commonly used in constituency-parsing evaluation. We additionally use \texttt{jp-evalb} because punctuation is retained as independent terminals in our converted Penn Korean trees. The punctuation-aware evaluation preserves distinctions that may be removed or normalized by standard \texttt{EVALB}, providing a stricter assessment of the predicted surface constituency structure. Reporting both metrics maintains comparability with previous constituency-parsing work while showing the effect of punctuation-sensitive evaluation.

\subsection{Overall results}
\label{sec:overall-results}

Table~\ref{tab:native-results} reports the native test-set results for the three representations. These scores evaluate each system in the representation in which it is trained and decoded: Morpheme+XPOS, Eojeol+XPOS, and Eojeol+UPOS. We report corpus-level labeled \(F_1\) under both evaluation settings.

\begin{table}
\centering
\footnotesize
\begin{tabular}{llrr}
\toprule
Representation & System & \texttt{EVALB} \(F_1\) & \texttt{jp-evalb} \(F_1\) \\
\midrule
Morpheme+XPOS    & Top-down  & 79.45 & 77.19 \\
    & In-order  & 83.13 & 80.68 \\
     & Bottom-up & 84.61 & 82.02 \\
\midrule
Eojeol+XPOS  & Top-down  & 69.62 & 67.73 \\
  & In-order  & 81.01 & 78.62 \\
  & Bottom-up & 80.80 & 78.48 \\
\midrule
Eojeol+UPOS  & Top-down  & 63.91 & 62.44 \\
  & In-order  & 69.97 & 68.19 \\
  & Bottom-up & 70.11 & 68.30 \\
\bottomrule
\end{tabular}
\caption{Native constituency parsing results on the test set. Each system is evaluated in its own output representation. 
}
\label{tab:native-results}
\end{table}

The native scores are informative, but they are not sufficient for a direct comparison between morpheme-based and eojeol-based parsing. The morpheme representation has a different terminal yield from the eojeol representations, and therefore defines a different set of possible spans. A higher or lower score in this setting may partly reflect the evaluation domain rather than only the quality of recovered eojeol-level phrase structure.

Table~\ref{tab:converted-results} reports a normalized comparison in the Eojeol+UPOS representation using the conversion procedure defined in Section~\ref{sec:normalization}. Morpheme+XPOS predictions are projected through the fixed morpheme-to-eojeol alignment, Eojeol+XPOS predictions retain their eojeol spans while their preterminals are replaced with UPOS labels, and native Eojeol+UPOS outputs remain unchanged. All conditions are therefore evaluated against the same gold terminal and preterminal representation.

\begin{table}
\centering
\footnotesize
\begin{tabular}{llrr}
\toprule
Representation & System & \texttt{EVALB} \(F_1\) & \texttt{jp-evalb} \(F_1\) \\
\midrule
Morpheme+XPOS \(\rightarrow\) Eojeol+UPOS & Top-down  & 79.58 & 77.32 \\
& In-order  & 83.26 & 80.79 \\
 & Bottom-up & 84.76 & 82.17 \\
\midrule
Eojeol+XPOS \(\rightarrow\) Eojeol+UPOS & Top-down  & 69.62 & 67.73 \\
 & In-order  & 81.01 & 78.62 \\
 & Bottom-up & 80.80 & 78.48 \\
\midrule
Eojeol+UPOS native & Top-down  & 63.91 & 62.44 \\
 & In-order  & 69.97 & 68.19 \\
 & Bottom-up & 70.11 & 68.30 \\
\bottomrule
\end{tabular}
\caption{Normalized constituency parsing results in the Eojeol+UPOS representation. Morpheme-based and Eojeol+XPOS predictions are converted to Eojeol+UPOS before evaluation, allowing all systems to be compared against the same gold representation. 
}
\label{tab:converted-results}
\end{table}

The normalized results show that the morpheme-based systems recover substantially more eojeol-level phrase structure than systems trained directly on Eojeol+UPOS trees. The strongest result is obtained by bottom-up parsing in the morpheme representation, reaching 84.76 \(F_1\) with \texttt{EVALB} and 82.17 \(F_1\) with \texttt{jp-evalb} after conversion to Eojeol+UPOS. The Eojeol+XPOS condition also improves strongly over Eojeol+UPOS, especially for in-order and bottom-up parsing, showing that fine-grained Korean XPOS information is much more useful than coarse UPOS labels even when evaluation is ultimately performed in the Eojeol+UPOS representation.\footnote{Previous parsing experiments on the Penn Korean Treebank include \citet{chung-post-gildea:2010:SPMRL}, who report up to 66.28 labeled \(F_1\) with a morpheme-based probabilistic spinal grammar, which is most directly comparable to the morpheme results in Table~\ref{tab:native-results}. Their model extracts spinal tree-substitution grammar fragments by maximally projecting each lexical item under manually specified Korean head rules, smooths the resulting fragment counts with depth-one treebank rules, and estimates rule probabilities by relative frequency \citep{post-gildea-2009-bayesian}.}

\section{Analysis}
\label{sec:analysis}

This section analyzes three structural properties of the parsing results. First, we compare the oracle transition sequences required by the three transition systems. This shows how the choice of transition order affects derivation length for the morpheme, Eojeol+UPOS, and Eojeol+XPOS representations. Second, we evaluate how well the predicted trees recover constituents of different arities after all outputs are converted to the unified Eojeol+UPOS representation. Third, we examine labeled $F_1$ by tree height, distinguishing local structures near the preterminal layer from higher-level constituents. All results in this section are computed with \texttt{jp-evalb}, using the punctuation-aware setting in which punctuation marks are included in the evaluated trees.

Table~\ref{tab:transition-statistics} reports transition statistics on the training split. Top-down and in-order parsing have the same oracle length, since both require one shift action per terminal and separate structural actions for opening and completing constituents. Bottom-up parsing yields much shorter derivations because its reduction actions directly construct completed constituents. The Morpheme+XPOS representation requires more transitions than the two eojeol representations because it has a longer terminal sequence. In contrast,  Eojeol+XPOS and Eojeol+UPOS have identical transition counts, since they share the same eojeol terminals and differ only in the preterminal label inventory.

\begin{table}
\centering
\footnotesize
\begin{tabular}{lrrrrrr}
\toprule
System
& M+XPOS avg.
& M+XPOS total
& E+XPOS avg.
& E+XPOS total
& E+UPOS avg.
& E+UPOS total \\
\midrule
Top-down & 217.88 & 844713 & 150.94 & 585201 & 150.94 & 585201 \\
In-order & 217.88 & 844713 & 150.94 & 585201 & 150.94 & 585201 \\
Bottom-up & 84.55 & 327798 & 62.24 & 241294 & 62.24 & 241294 \\
\bottomrule
\end{tabular}
\caption{Transition statistics on the training split. The counts are computed from implemented gold-oracle transition sequences and exclude \textsc{Finish}. M+XPOS, E+UPOS, and E+XPOS denote the Morpheme+XPOS, Eojeol+XPOS, and Eojeol+UPOS representations, respectively.}
\label{tab:transition-statistics}
\end{table}

Table~\ref{tab:high-arity-f1} reports arity-specific labeled \(F_1\) on the test set after conversion to Eojeol+UPOS. This evaluation asks whether each system recovers the same eojeol-level constituent structure, including unary, binary, and higher-arity constituents. The comparison is therefore not affected by differences in terminal granularity or preterminal label sets. The results show that the morpheme-based systems consistently recover higher-quality eojeol-level structure than the native Eojeol+UPOS systems, especially for binary and higher-arity constituents. The Eojeol+XPOS systems also improve substantially over Eojeol+UPOS, indicating that fine-grained Korean XPOS information is useful even when the final evaluation is performed with UPOS preterminals.

\begin{table}
\centering
\footnotesize
\begin{tabular}{llrrrrr}
\toprule
Representation
& System
& \(k=1\)
& \(k=2\)
& \(k=3\)
& \(k=4\)
& \(k \geq 5\) \\
\midrule
Morpheme+XPOS \(\rightarrow\) Eojeol+UPOS
& Top-down & 82.90 & 70.45 & 66.89 & 54.03 & 59.04 \\
Eojeol+XPOS \(\rightarrow\) Eojeol+UPOS
&  & 70.67 & 59.57 & 59.56 & 48.68 & 57.46 \\
Eojeol+UPOS native
&  & 68.80 & 48.85 & 54.27 & 39.74 & 57.14 \\ 
\midrule

Morpheme+XPOS \(\rightarrow\) Eojeol+UPOS
& In-order & 86.49 & 75.14 & 72.70 & 58.46 & 67.22 \\
Eojeol+XPOS \(\rightarrow\) Eojeol+UPOS
&  & 84.58 & 70.77 & 71.17 & 59.83 & 67.34 \\
Eojeol+UPOS native
&  & 74.07 & 57.21 & 64.12 & 46.90 & 59.55 \\
\midrule

Morpheme+XPOS \(\rightarrow\) Eojeol+UPOS
& Bottom-up & 87.52 & 77.12 & 74.75 & 63.83 & 69.88 \\
Eojeol+XPOS \(\rightarrow\) Eojeol+UPOS
&  & 84.11 & 70.94 & 70.50 & 59.57 & 69.12 \\
Eojeol+UPOS native
&  & 74.96 & 56.88 & 62.83 & 47.79 & 60.08 \\

\bottomrule
\end{tabular}
\caption{Arity-specific constituent results on the test split in the unified Eojeol+UPOS representation. Native Eojeol+UPOS predictions are evaluated directly, while Eojeol+XPOS and morpheme-based predictions are converted to Eojeol+UPOS before evaluation.}
\label{tab:high-arity-f1}
\end{table}

Across transition systems, bottom-up parsing gives the strongest arity-specific results after morpheme-to-eojeol conversion, with the highest \(F_1\) for unary, binary, ternary, four-child, and high-arity constituents. In-order parsing is generally close to bottom-up parsing, especially in the Eojeol+XPOS and morpheme-converted settings. Top-down parsing is weaker across most arities, but it shows the same overall pattern: richer morphological or XPOS information improves recovery of eojeol-level phrase structure.

Table~\ref{tab:f1-by-height} further breaks down labeled $F_1$ by tree height, where terminals have height 0 and preterminals have height 1. Since POS labels are provided as input, the height-1 scores are essentially perfect for all systems. The largest representation effect appears immediately above the preterminal layer, at height 2: for example, under top-down parsing, morpheme-based prediction reaches 91.09 $F_1$, compared with 86.36 for Eojeol+XPOS and 78.69 for native Eojeol+UPOS. This supports the view that the main advantage of the morpheme-based representation comes from detailed morphosyntactic information that improves local phrase-structure decisions. However, the advantage does not disappear at higher levels. In the top-down system, the morpheme-based representation remains substantially better than the two eojeol-based alternatives across all heights, including larger constituents. Thus, better local structural decisions are not merely localized effects; they also propagate upward and improve the construction of higher-level constituents.


\begin{table}[!t]
\centering
\footnotesize
\setlength{\tabcolsep}{3.5pt}
\begin{tabular}{llrrrrrrrr}
\toprule
Representation & System
& $h{=}1$ & $h{=}2$ & $h{=}3$ & $h{=}4$
& $h{=}5$ & $h{=}6$ & $h{=}7$ & $h{\geq}8$ \\
\midrule
Morpheme+XPOS $\rightarrow$ Eojeol+UPOS & Top-down
& 100.00 & 91.09 & 77.37 & 67.98 & 58.95 & 49.66 & 42.47 & 22.94 \\
Eojeol+XPOS $\rightarrow$ Eojeol+UPOS & 
& 100.00 & 86.36 & 66.45 & 55.27 & 47.96 & 39.33 & 31.83 & 13.43 \\
Eojeol+UPOS native& 
& 100.00 & 78.69 & 50.34 & 42.38 & 35.67 & 28.26 & 19.17 & 11.60 \\
\midrule
Morpheme+XPOS $\rightarrow$ Eojeol+UPOS & In-order
& 100.00 & 92.19 & 81.03 & 72.57 & 65.60 & 56.93 & 50.25 & 33.49 \\
Eojeol+XPOS $\rightarrow$ Eojeol+UPOS & 
& 100.00 & 91.33 & 73.78 & 65.67 & 59.17 & 52.84 & 45.19 & 28.85 \\
Eojeol+UPOS native& 
& 100.00 & 81.17 & 57.59 & 50.01 & 44.18 & 35.33 & 26.50 & 19.12 \\
\midrule
Morpheme+XPOS $\rightarrow$ Eojeol+UPOS & Bottom-up
& 100.00 & 92.42 & 81.92 & 74.08 & 66.93 & 60.53 & 52.65 & 35.24 \\
Eojeol+XPOS $\rightarrow$ Eojeol+UPOS & 
& 100.00 & 91.16 & 74.61 & 67.17 & 58.05 & 50.57 & 43.24 & 30.69 \\
Eojeol+UPOS native& 
& 100.00 & 81.49 & 57.83 & 50.02 & 43.21 & 36.79 & 28.89 & 18.84 \\
\bottomrule
\end{tabular}
\caption{Results by tree height. Terminals have height 0, and preterminals have height 1. All systems are evaluated after normalization to Eojeol+UPOS trees.}
\label{tab:f1-by-height}
\end{table}

The arity- and height-based analyses suggest that morpheme-based parsing provides useful local morphosyntactic evidence for recovering eojeol-level phrase structure, and that this advantage extends beyond the immediate preterminal region. We return to this point in Section~\ref{sec:discussion-morpheme-eojeol}.

\section{Discussion}
\label{sec:discussion}

This paper addresses a representational question central to Korean constituency annotation: what should count as the terminal unit of a phrase-structure tree? For Korean, this is not a simple tokenization choice. An eojeol is the ordinary spacing unit of written Korean, but it is internally complex, often combining lexical material with case particles, verbal endings, derivational suffixes, auxiliaries, or other functional morphemes \citep{park-kim-2023-role}. A morpheme-based tree projects this internal composition into the constituency yield, whereas an eojeol-based tree treats the same material as part of a surface syntactic unit. The two representations therefore differ not only in terminal granularity, but also in how morphology is related to phrase structure.

Two conclusions must be distinguished. Empirically, under gold segmentation and preterminal annotation, the evaluated transition-based parsers recover more eojeol-level constituency structure in the Morpheme+XPOS and Eojeol+XPOS conditions than in Eojeol+UPOS. This finding establishes the value of fine-grained morphosyntactic information under the evaluated parsing conditions, but it does not independently determine which units should constitute the terminal domain of a Korean phrase-structure treebank. The argument for eojeol as that domain is instead linguistic and representational: it concerns the separation of phrase-level grouping from eojeol-internal morphology, the overt character of spacing-unit boundaries, and compatibility with other Korean syntactic resources.

The position defended here is that eojeols provide a consistent and linguistically transparent terminal domain for Korean constituency treebank design, even though morpheme-level information provides stronger parsing cues under the evaluated conditions. This does not mean that Korean syntax is independent of morphology. On the contrary, functional morphology is central to syntactic interpretation in Korean. Case particles, topic markers, connective endings, adnominal endings, verbal inflection, and derivational morphology all provide information relevant to argument structure, modification, subordination, and clause type. The question is whether these morphemes should be represented as independent constituency terminals. Following the synthetic view of Korean wordhood, in which bound functional morphemes are grammatically integrated within a larger syntactic unit rather than functioning as independent syntactic words \citep{chulwoo-2019-for-developing}, we argue that constituency structure can be coherently defined over eojeol terminals while morpheme-level information is preserved in an aligned annotation layer.

Our eojeol-based conversion of the Penn Korean Treebank makes this separation explicit. Phrase structure is defined over overt eojeol terminals, while morpheme sequences and fine-grained Korean POS tags remain available as aligned, non-constituent annotation. This design represents syntactic grouping over the surface units of Korean text without discarding the morphosyntactic information needed for analysis or parsing. It also aligns constituency annotation with Korean dependency resources, in which syntactic relations are normally defined over eojeol-level tokens \citep{noh-etal-2018-enhancing,seo-etal-2019-ud,han-etal-2020-annotation,park-etal-2021-klue,kim-etal-2024-kud}.

\subsection{Eojeol as a constituency terminal domain}
\label{sec:discussion-eojeol}


The linguistic motivation for eojeol-based constituency is that {constituency annotation should encode phrase-level grouping, not word-internal morphological composition}. Korean functional morphology is syntactically consequential, but its contribution is typically mediated through the eojeol in which it appears. A case-marked nominal eojeol, for example, participates in argument structure as a surface syntactic unit, even though its internal morphemes identify its grammatical role. Similarly, {verbal endings may mark clausal embedding, modification, coordination, tense, mood, or modality, but these features are expressed inside the verbal eojeol} rather than as independent surface words.

Projecting each morpheme as a constituency terminal therefore changes the representational object. The tree no longer directly represents phrase structure over the surface sentence, but over a morphologically segmented sequence. This may be appropriate for theoretical analyses that explicitly project functional morphology, but it introduces additional commitments into a treebank whose goal is to provide a stable surface constituency backbone. Morphological attachment patterns may then appear as constituency boundaries, and category assignments for bound morphemes may be interpreted as phrase-structural distinctions even when they primarily reflect word-internal composition.

The eojeol-based representation avoids this conflation. Constituency labels and hierarchical relations are interpreted as relations among surface syntactic units, while morphology is retained as an aligned layer. Under this view, a form such as 영국정부는 \textit{yeonggukjeongbuneun} (`the British government-\textsc{top}') is a single terminal in the constituency tree, but its internal analysis remains available for interpretation and parser input. This gives the output tree a direct surface interpretation while preserving access to the morphology that makes Korean syntax recoverable.


This distinction is also important for evaluation. If constituency trees are morpheme-based, phrase boundaries are evaluated over a sequence that is not directly present as the spacing-unit sequence of the sentence. If trees are eojeol-based, every constituent span corresponds to a span over overt eojeol. This makes the representation better aligned with downstream processing and with eojeol-based dependency treebanks \citep{park:2017:Depling,park-kim-2024-word}. The point is not that morpheme-based annotation is uninformative, but that morpheme segmentation should not itself determine the terminal domain of constituency evaluation.

{An eojeol-based representation is compatible with accounts of the Korean morphology-syntax interface. \citet{cho-sells-1995} argue that Korean nominal and verbal inflectional suffixes combine with their hosts in the lexical component, although the resulting forms contribute information required by syntax. In their HPSG-based Korean Phrase Structure Grammar, \citet{kim-yang-2003,kim-yang-2004} similarly use inflectional and lexical rules to construct fully inflected word signs before phrasal combination, with information such as case and verbal inflection represented in typed feature structures rather than by projecting each bound exponent as an independent syntactic daughter. \citet{park-park-2026-constituency} likewise argue for an eojeol-based constituency backbone across Korean treebanks, with morphological segmentation and fine-grained POS information represented in a parallel, non-constituent layer. These analyses do not entail that every orthographic eojeol is categorically identical to a theoretical grammatical word; they support a more limited but important architectural claim that bound morphology need not occupy the same representational layer as phrase structure. Combined with the overt and reproducible boundaries supplied by Korean spacing, this provides a formal motivation for using eojeols as the primary treebank terminals.}

\subsection{Morpheme-based and eojeol-based parsing}
\label{sec:discussion-morpheme-eojeol}

The empirical results show that morpheme-based parsing performs best even when its predictions are projected back to the Eojeol+UPOS representation. This result is consistent with previous Korean NLP work. Morpheme-based annotation has been shown to be effective for POS tagging \citep{park-tyers:2019:LAW}, dependency parsing \citep{chen-etal-2022-yet}, named entity recognition \citep{chen-lim-park-2024-korean}, and Korean FrameNet annotation \citep{chen-etal-2024-towards-standardized}. Across these tasks, morpheme-level annotation provides fine-grained cues that are often hidden within eojeol surface forms.

Our parsing results follow this broader pattern. In the morpheme-based representation, particles, endings, derivational suffixes, and auxiliary material are exposed as separate terminals with XPOS preterminals. These units provide direct evidence for local phrase-label decisions. Case particles and topic markers help identify nominal functions; connective and adnominal endings signal clausal modification or subordination; and verbal endings help distinguish clause types and predicate environments. When these cues are explicit in the input and terminal representation, they can improve local structural decisions, which may in turn support the recovery of larger constituents.

Under the gold-annotation conditions used here, this empirical advantage should not be interpreted as a representational argument that morphemes must be constituency terminals.
The projected evaluation shows that a morpheme-based representation can improve the recovery of eojeol-level phrase structure. The results therefore do not imply that Korean constituency is inherently morpheme-based; rather, they show that morpheme-level information provides valuable evidence for recovering constituency over eojeols.

The Eojeol+XPOS results reinforce this interpretation. Compared with Eojeol+UPOS, Eojeol+XPOS substantially improves parsing while preserving the same eojeol terminal yield, showing that the performance difference is not attributable only to terminal-sequence length. Fine-grained Korean XPOS information provides valuable evidence for phrase-structure prediction within an eojeol-based terminal domain. 
The linguistic and resource-design considerations developed above motivate separating the treebank-design question from the information available for parsing: eojeol can serve as the constituency terminal domain, while the experimental results show that morpheme-level annotation and XPOS labels provide valuable morphosyntactic evidence.
The practical recommendation is therefore to keep constituency output eojeol-based while preserving morpheme sequences and XPOS information as aligned annotation that provides valuable evidence for parsing.

\subsection{Transition order and Korean phrase structure}
\label{sec:discussion-transition-order}

The comparison of top-down, in-order, and bottom-up systems shows how transition order affects phrase-structure prediction. Top-down parsing predicts a phrase label before its children are available; in-order parsing predicts the label after observing the left corner; and bottom-up parsing waits until the complete child sequence has been constructed. Although the systems derive the same class of non-binary trees, they differ in how much structural evidence is available when the mother label is predicted.

Bottom-up parsing generally performs best in our experiments, with in-order parsing often close and top-down parsing weaker. This ordering is not specific to Korean: we observe a similar advantage for bottom-up parsing in corresponding experiments on the English Penn Treebank and Chinese Treebank [CITATION OR CROSS-REFERENCE]. The Korean results are therefore consistent with a broader benefit of delaying phrase-label prediction until more of the local constituent structure is available.

Korean morphosyntax may further reinforce this benefit because important syntactic cues frequently occur at the right edge of an eojeol or clause. Particles mark the grammatical roles of nominal expressions, while verbal endings indicate clausal relations, modification, coordination, and clause type. By delaying the mother-label decision, bottom-up parsing can condition that decision on the complete child sequence, including right-edge morphosyntactic evidence. This is consistent with its strong performance in the Morpheme+XPOS and Eojeol+XPOS conditions, although the parallel English and Chinese results indicate that right-edge Korean morphology is not the sole source of the bottom-up advantage.

The arity-specific results further show that the benefit of richer morphosyntactic information is not limited to unary or binary constituents. Improvements also occur for ternary, four-child, and higher-arity constituents. This is relevant because many structures in the converted treebank are non-binary, including nominal sequences, genitive and adnominal modifiers, adjunct sequences, and clausal projections. Direct non-binary transition systems can recover these structures without introducing artificial intermediate nodes.

\subsection{Relation to dependency resources}
\label{sec:discussion-dependency}

An eojeol-based constituency representation also clarifies the relation between Korean constituency and dependency annotation. Korean dependency resources are predominantly organized over eojeol-level tokens \citep{noh-etal-2018-enhancing,seo-etal-2019-ud,han-etal-2020-annotation,park-etal-2021-klue,kim-etal-2024-kud}. If constituency trees use the same terminal domain, then constituency and dependency analyses can be compared without first reconstructing word boundaries from morpheme sequences.

This does not collapse the distinction between the two formalisms. Dependency trees encode labeled head-dependent relations, while constituency trees encode hierarchical grouping. The value of eojeol-based constituency is that it makes the two representations commensurable at the level of the surface yield. The same eojeol sequence can support a dependency analysis and a constituency analysis, while morphology remains available as an aligned layer for both. This shared terminal domain is essential for constituency--dependency conversion, head-finding, and cross-formalism evaluation.\footnote{{Because the systems are trained under different terminal and preterminal representations, a conventional error analysis would conflate parser errors with representation-induced differences. We therefore analyze the outputs after conversion to a unified Eojeol+UPOS representation, focusing on arity-specific and height-specific labeled F1. These analyses directly measure how terminal granularity and morphosyntactic category granularity affect the recovery of eojeol-level constituency structure.}}

\subsection{Limitations}
\label{sec:discussion-limitations}

The scope of this paper is limited to the Penn Korean Treebank. This makes the conversion relatively conservative, because the original resource already contains eojeol-level surface units with internal morphological annotation. The main conversion steps are null-element removal, alignment with the eojeol sequence, UPOS preterminal assignment, and extraction of morpheme-based and eojeol-based targets. The resulting resource should therefore be understood as one principled eojeol-based conversion of the Penn Korean Treebank, not as a complete normalization of all Korean constituency resources.

Other Korean treebanks raise different issues. A Sejong conversion would require principled debinarization, while a KAIST conversion would require decisions about how morpheme-level and function-morpheme projections should be collapsed into eojeol terminals. These are not merely technical choices. They determine which syntactic distinctions are preserved, flattened, or reassigned when moving from a morpheme-sensitive representation to an eojeol-based constituency backbone.

A second limitation concerns null elements. Removing null elements is appropriate for surface-oriented parsing, but it removes part of the analysis encoded in the original Penn Korean Treebank. The converted trees do not ask the parser to recover empty subjects, traces, empty operators, or predicate deletions. They therefore define a surface constituency parsing task, not a full reconstruction of the original empty-category analysis.

A further limitation concerns the relationship between the evaluated representations and the proposed separation of constituency output from morphosyntactic evidence. Morpheme+XPOS changes the terminal sequence and parsing target, whereas Eojeol+XPOS supplies combined gold XPOS sequences as preterminal labels. Moreover, the normalized evaluation uses gold eojeol boundaries and the fixed gold morpheme-to-eojeol alignment when projecting Morpheme+XPOS predictions; it therefore measures phrase-structure recovery in a common eojeol span domain rather than the parser’s ability to recover eojeol boundaries or morphological alignment. The experiments motivate, but do not instantiate, an eojeol-output parser that uses aligned morpheme sequences and XPOS labels as auxiliary features. Constituency parsing is conventionally evaluated over a given token sequence, and joint prediction of morphological segmentation, POS annotation, and constituency structure is less commonly treated as part of the constituency parsing task than in dependency-parsing pipelines. Nevertheless, evaluating the proposed configuration with predicted morphological annotation would provide a useful extension toward an end-to-end Korean constituency parsing setting.

Finally, the results depend on the conversion and evaluation choices adopted here, including punctuation handling, UPOS mapping, functional label treatment, unary-chain handling after null-element deletion, and projection of morpheme-based predictions back to the eojeol domain. Wherever possible, we follow the general preprocessing conventions used for Penn Treebank-style constituency parsing, including the removal of null elements and the normalization of structures for labeled-bracketing evaluation. These choices are conservative and reproducible, but they define a specific parsing target. Future work should test how robust the findings are under alternative conversion policies and extend the comparison to other Korean constituency resources.

\section{Conclusion}
\label{sec:conclusion}

{In this paper, we argue for an eojeol-based constituency treebank representation for Korean based on the Penn Korean Treebank}. The conversion removes null elements, aligns Penn Korean phrase structure with overt eojeol tokens, assigns UPOS preterminals, and preserves the original phrase-structural analysis where possible. The resulting trees define constituency over the surface eojeol sequence, while keeping morpheme sequences and fine-grained Korean POS tags as aligned morphosyntactic annotation.

The experiments show that terminal granularity has a substantial effect on Korean constituency parsing. Native morpheme-based scores, native eojeol-based scores, and converted eojeol-domain scores are not interchangeable, because morpheme and eojeol representations define different terminal yields and therefore different span spaces. 
By projecting Morpheme+XPOS and Eojeol+XPOS predictions into the unified Eojeol+UPOS representation, we place all predictions in a common eojeol span domain and compare how much eojeol-level phrase structure each representation recovers under its respective gold input condition.

The strongest results come from Morpheme+XPOS parsing after projection to Eojeol+UPOS, followed by Eojeol+XPOS parsing. Under the gold segmentation and preterminal conditions used here, this pattern shows that fine-grained morphological and XPOS representations provide valuable evidence for the evaluated transition-based parsers. It does not, by itself, establish whether morphemes or eojeols are the linguistically preferable constituency terminals.

Our preference for an eojeol terminal domain instead follows from the linguistic and resource-design argument developed in this paper: eojeols provide overt surface boundaries and a shared domain for alignment with other Korean resources, while morpheme sequences and XPOS labels can remain available as aligned morphosyntactic annotation.

Across transition systems, bottom-up and in-order parsing generally outperform top-down parsing, especially when rich morphological information is available. The arity-based analysis further shows that these gains extend beyond binary constituents to ternary, four-child, and higher-arity structures. This supports the use of direct non-binary transition systems for Korean constituency parsing, where many nominal, modifier, adjunct, and clausal structures are naturally represented without artificial binarization.

These results provide a conservative foundation for surface-oriented Korean constituency parsing and for future work on Korean treebank conversion, constituency--dependency interoperability, and parsing models that exploit morphology while maintaining eojeol as the terminal domain of phrase structure. 
More broadly, they motivate a surface-oriented design for Korean constituency resources in which eojeol define the phrase-structure output domain while morpheme-level and XPOS information remain available as aligned morphosyntactic evidence.

\paragraph{Acknowledgments}
We thank the Stanza developers for making the Stanza constituency parser architecture publicly available; our implementation is based on this framework.


\begin{thebibliography}{54}
\providecommand{\natexlab}[1]{#1}
\providecommand{\url}[1]{\texttt{#1}}
\expandafter\ifx\csname urlstyle\endcsname\relax
  \providecommand{\doi}[1]{doi: #1}\else
  \providecommand{\doi}{doi: \begingroup \urlstyle{rm}\Url}\fi

\bibitem[Abeill{\'{e}}(2003)]{abeille:2003}
Anne Abeill{\'{e}}, editor.
\newblock \emph{{Treebanks: Building and Using Parsed Corpora}}.
\newblock Springer Netherlands, 2003.
\newblock ISBN 978-1-4020-1334-8.
\newblock \doi{10.1007/978-94-010-0201-1}.

\bibitem[Bangalore and Joshi(1999)]{bangalore-joshi:1999:CL}
Srinivas Bangalore and Aravind~K Joshi.
\newblock {Supertagging: An Approach to Almost Parsing}.
\newblock \emph{Computational Linguistics}, 25\penalty0 (2):\penalty0 237--265, 1999.

\bibitem[Bauer and Manning(2025)]{bauer-manning-2025-high}
John Bauer and Christopher~D. Manning.
\newblock {High-Accuracy Transition-Based Constituency Parsing}.
\newblock In Kenji Sagae and Stephan Oepen, editors, \emph{Proceedings of the 18th International Conference on Parsing Technologies (IWPT, SyntaxFest 2025)}, pages 26--39, Ljubljana, Slovenia, 8 2025. Association for Computational Linguistics.
\newblock ISBN 979-8-89176-294-7.
\newblock URL \url{https://aclanthology.org/2025.iwpt-1.4/}.

\bibitem[Bj{\"{o}}rkelund et~al.(2013)Bj{\"{o}}rkelund, {\c{C}}etino{\u{g}}lu, Farkas, Mueller, and Seeker]{bjorkelund-EtAl:2013:SPMRL}
Anders Bj{\"{o}}rkelund, Özlem {\c{C}}etino{\u{g}}lu, Richárd Farkas, Thomas Mueller, and Wolfgang Seeker.
\newblock {(Re)ranking Meets Morphosyntax: State-of-the-art Results from the SPMRL 2013 Shared Task}.
\newblock In \emph{Proceedings of the Fourth Workshop on Statistical Parsing of Morphologically-Rich Languages}, pages 135--145, Seattle, Washington, USA, 10 2013. Association for Computational Linguistics.
\newblock URL \url{https://www.aclweb.org/anthology/W13-4916}.

\bibitem[Bj{\"{o}}rkelund et~al.(2014)Bj{\"{o}}rkelund, {\c{C}}etino{\u{g}}lu, Fale{\'{n}}ska, Farkas, Mueller, Seeker, and Sz{\'{a}}nt{\'{o}}]{bjorkelund-EtAl:2014:SPMRL}
Anders Bj{\"{o}}rkelund, Özlem {\c{C}}etino{\u{g}}lu, Agnieszka Fale{\'{n}}ska, Richárd Farkas, Thomas Mueller, Wolfgang Seeker, and Zsolt Sz{\'{a}}nt{\'{o}}.
\newblock {Introducing the IMS-Wroc{\l}aw-Szeged-CIS entry at the SPMRL 2014 Shared Task: Reranking and Morpho-syntax meet Unlabeled Data}.
\newblock In \emph{Proceedings of the First Joint Workshop on Statistical Parsing of Morphologically Rich Languages and Syntactic Analysis of Non-Canonical Languages}, pages 97--102, Dublin, Ireland, 8 2014. Dublin City University.
\newblock URL \url{https://www.aclweb.org/anthology/W14-6110}.

\bibitem[Black et~al.(1991)Black, Abney, Flickinger, Gdaniec, Grishman, Harrison, Hindle, Ingria, Jelinek, Klavans, Liberman, Marcus, Roukos, Santorini, and Strzalkowski]{black-etal-1991-procedure}
Ezra Black, Steven~P. Abney, Dan Flickinger, Claudia Gdaniec, Ralph Grishman, Phil Harrison, Don Hindle, Robert Ingria, Frederick Jelinek, Judith Klavans, Mark Liberman, Mitch Marcus, Salim Roukos, Beatrice Santorini, and Tomek Strzalkowski.
\newblock {A Procedure for Quantitatively Comparing the Syntactic Coverage of English Grammars}.
\newblock In \emph{Proceedings of a Workshop on Speech and Natural Language}, pages 306--311, Pacific Grove, California, 2 1991. Morgan Kaufmann.
\newblock URL \url{https://aclanthology.org/H91-1060/}.

\bibitem[Chen et~al.(2022)Chen, Jo, Yao, Lim, Silfverberg, Tyers, and Park]{chen-etal-2022-yet}
Yige Chen, Eunkyul~Leah Jo, Yundong Yao, KyungTae Lim, Miikka Silfverberg, Francis~M. Tyers, and Jungyeul Park.
\newblock {Yet Another Format of Universal Dependencies for Korean}.
\newblock In Nicoletta Calzolari, Chu-Ren Huang, Hansaem Kim, James Pustejovsky, Leo Wanner, Key-Sun Choi, Pum-Mo Ryu, Hsin-Hsi Chen, Lucia Donatelli, Heng Ji, Sadao Kurohashi, Patrizia Paggio, Nianwen Xue, Seokhwan Kim, Younggyun Hahm, Zhong He, Tony~Kyungil Lee, Enrico Santus, Francis Bond, and Seung-Hoon Na, editors, \emph{Proceedings of the 29th International Conference on Computational Linguistics}, pages 5432--5437, Gyeongju, Republic of Korea, 10 2022. International Committee on Computational Linguistics.
\newblock URL \url{https://aclanthology.org/2022.coling-1.482}.

\bibitem[Chen et~al.(2024{\natexlab{a}})Chen, Ihn, Lim, and Park]{chen-etal-2024-towards-standardized}
Yige Chen, Jae Ihn, KyungTae Lim, and Jungyeul Park.
\newblock {Towards Standardized Annotation and Parsing for Korean FrameNet}.
\newblock In Nicoletta Calzolari, Min-Yen Kan, Veronique Hoste, Alessandro Lenci, Sakriani Sakti, and Nianwen Xue, editors, \emph{Proceedings of the 2024 Joint International Conference on Computational Linguistics, Language Resources and Evaluation (LREC-COLING 2024)}, pages 16653--16658, Torino, Italy, 5 2024{\natexlab{a}}. ELRA and ICCL.
\newblock URL \url{https://aclanthology.org/2024.lrec-main.1447}.

\bibitem[Chen et~al.(2024{\natexlab{b}})Chen, Lim, and Park]{chen-lim-park-2024-korean}
Yige Chen, KyungTae Lim, and Jungyeul Park.
\newblock {Korean named entity recognition based on language-specific features}.
\newblock \emph{Natural Language Engineering}, 30\penalty0 (3):\penalty0 625–649, 2024{\natexlab{b}}.
\newblock \doi{10.1017/S1351324923000311}.

\bibitem[Choi et~al.(2012)Choi, Park, and Choi]{choi-park-choi:2012:SP-SEM-MRL}
DongHyun Choi, Jungyeul Park, and Key-Sun Choi.
\newblock {Korean Treebank Transformation for Parser Training}.
\newblock In \emph{Proceedings of the ACL 2012 Joint Workshop on Statistical Parsing and Semantic Processing of Morphologically Rich Languages}, pages 78--88, Jeju, Republic of Korea, 2012. Association for Computational Linguistics.
\newblock URL \url{http://www.aclweb.org/anthology/W12-3411}.

\bibitem[Choi et~al.(1994)Choi, Han, Han, and Kwon]{choi-EtAl:1994}
Key-Sun Choi, Young~S. Han, Young~G. Han, and Oh~W. Kwon.
\newblock {KAIST Tree Bank Project for Korean: Present and Future Development}.
\newblock In \emph{Proceedings of the International Workshop on Sharable Natural Language Resources}, pages 7--14, Nara Institute of Science and Technology, 1994. Nara Institute of Science and Technology.

\bibitem[Chomsky(1981)]{chomsky:1981}
Noam Chomsky.
\newblock \emph{{Lectures on Government and Binding}}.
\newblock Studies in Generative Grammar. Foris Publications, Dordrecht, The Netherlands, 1981.

\bibitem[Chomsky(1986)]{chomsky-1986-barriers}
Noam Chomsky.
\newblock \emph{{Barriers}}.
\newblock Linguistic Inquiry Monograph 13. The MIT Press, Cambridge, MA, 1986.
\newblock ISBN 9780262031189.

\bibitem[Chun et~al.(2018)Chun, Han, Hwang, and Choi]{chun-EtAl:2018:LREC}
Jayeol Chun, Na-Rae Han, Jena~D. Hwang, and Jinho~D. Choi.
\newblock {Building Universal Dependency Treebanks in Korean}.
\newblock In \emph{Proceedings of the Eleventh International Conference on Language Resources and Evaluation (LREC 2018)}, Miyazaki, Japan, 2018. European Language Resources Association (ELRA).
\newblock ISBN 979-10-95546-00-9.

\bibitem[Chung et~al.(2010)Chung, Post, and Gildea]{chung-post-gildea:2010:SPMRL}
Tagyoung Chung, Matt Post, and Daniel Gildea.
\newblock {Factors Affecting the Accuracy of Korean Parsing}.
\newblock In \emph{Proceedings of the NAACL HLT 2010 First Workshop on Statistical Parsing of Morphologically-Rich Languages}, pages 49--57, Los Angeles, CA, USA, 2010. Association for Computational Linguistics.
\newblock URL \url{http://www.aclweb.org/anthology/W10-1406}.

\bibitem[Dyer et~al.(2016)Dyer, Kuncoro, Ballesteros, and Smith]{dyer-etal-2016-recurrent}
Chris Dyer, Adhiguna Kuncoro, Miguel Ballesteros, and Noah~A. Smith.
\newblock {Recurrent Neural Network Grammars}.
\newblock In Kevin Knight, Ani Nenkova, and Owen Rambow, editors, \emph{Proceedings of the 2016 Conference of the North American Chapter of the Association for Computational Linguistics: Human Language Technologies}, pages 199--209, San Diego, California, 6 2016. Association for Computational Linguistics.
\newblock \doi{10.18653/v1/N16-1024}.
\newblock URL \url{https://aclanthology.org/N16-1024}.

\bibitem[Fern{\'{a}}ndez-Gonz{\'{a}}lez and G{\'{o}}mez-Rodr{\'{i}}guez(2019)]{gonzalez-rodriguez-2019-faster}
Daniel Fern{\'{a}}ndez-Gonz{\'{a}}lez and Carlos G{\'{o}}mez-Rodr{\'{i}}guez.
\newblock {Faster shift-reduce constituent parsing with a non-binary, bottom-up strategy}.
\newblock \emph{Artificial Intelligence}, 275:\penalty0 559--574, 2019.
\newblock ISSN 0004-3702.
\newblock \doi{https://doi.org/10.1016/j.artint.2019.07.006}.
\newblock URL \url{https://www.sciencedirect.com/science/article/pii/S000437021830540X}.

\bibitem[Fried et~al.(2019)Fried, Kitaev, and Klein]{fried-nikita-klein:2019:ACL}
Daniel Fried, Nikita Kitaev, and Dan Klein.
\newblock {Cross-Domain Generalization of Neural Constituency Parsers}.
\newblock In \emph{Proceedings of the 57th Annual Meeting of the Association for Computational Linguistics}, pages 323--330, Florence, Italy, 7 2019. Association for Computational Linguistics.
\newblock URL \url{https://www.aclweb.org/anthology/P19-1031}.

\bibitem[Hahm et~al.(2018)Hahm, Kim, Kwon, and Choi]{hahm-EtAl:2018:LREC}
Younggyun Hahm, Jiseong Kim, Sunggoo Kwon, and Key-Sun Choi.
\newblock {Semi-automatic Korean FrameNet Annotation over KAIST Treebank}.
\newblock In Nicoletta Calzolari, Khalid Choukri, Christopher Cieri, Thierry Declerck, Sara Goggi, Koiti Hasida, Hitoshi Isahara, Bente Maegaard, Joseph Mariani, Hélène Mazo, Asuncion Moreno, Jan Odijk, Stelios Piperidis, and Takenobu Tokunaga, editors, \emph{Proceedings of the Eleventh International Conference on Language Resources and Evaluation (LREC 2018)}, Miyazaki, Japan, 5 2018. European Language Resources Association (ELRA).
\newblock ISBN 979-10-95546-00-9.

\bibitem[Han et~al.(2001)Han, Han, and Ko]{han-han-ko:2001}
Chung-Hye Han, Na-Rae Han, and Eon-Suk Ko.
\newblock {Bracketing Guidelines for Penn Korean TreeBank}.
\newblock Technical report, University of Pennsylvania, 2001.
\newblock URL \url{ftp://ftp.cis.upenn.edu/pub/ircs/tr/01-10/}.

\bibitem[Han et~al.(2002)Han, Han, Ko, Palmer, and Yi]{han-EtAl:2002}
Chung-Hye Han, Na-Rae Han, Eon-Suk Ko, Martha Palmer, and Heejong Yi.
\newblock {Penn Korean Treebank: Development and Evaluation}.
\newblock In \emph{Proceedings of the 16th Pacific Asia Conference on Language, Information and Computation}, pages 69--78, Jeju, Korea, 2002. Pacific Asia Conference on Language, Information and Computation.

\bibitem[Han et~al.(2020)Han, Oh, Jin, and Kim]{han-etal-2020-annotation}
Ji~Yoon Han, Tae~Hwan Oh, Lee Jin, and Hansaem Kim.
\newblock {Annotation Issues in Universal Dependencies for Korean and Japanese}.
\newblock In Marie-Catherine de~Marneffe, Miryam de~Lhoneux, Joakim Nivre, and Sebastian Schuster, editors, \emph{Proceedings of the Fourth Workshop on Universal Dependencies (UDW 2020)}, pages 99--108, Barcelona, Spain (Online), 12 2020. Association for Computational Linguistics.
\newblock URL \url{https://aclanthology.org/2020.udw-1.12}.

\bibitem[Hermjakob(2000)]{hermjakob:2000:NAACL}
Ulf Hermjakob.
\newblock {Rapid Parser Development: A Machine Learning Approach for Korean}.
\newblock In \emph{Proceedings of the 1st Meeting of the North American Chapter of the Association for Computational Linguistics}, pages 118--123, Seattle, Washington, USA, 2000.

\bibitem[Hermjakob and Mooney(1997)]{hermjakob-mooney:1997:EACL}
Ulf Hermjakob and Raymond~J. Mooney.
\newblock {Learning Parse and Translation Decisions from Examples with Rich Context}.
\newblock In \emph{35th Annual Meeting of the Association for Computational Linguistics and 8th Conference of the European Chapter of the Association for Computational Linguistics}, pages 482--489, Madrid, Spain, 7 1997. Association for Computational Linguistics.
\newblock \doi{10.3115/976909.979679}.
\newblock URL \url{https://www.aclweb.org/anthology/P97-1062}.

\bibitem[Jo et~al.(2024)Jo, Park, and Park]{jo-etal-2024-novel}
Eunkyul~Leah Jo, Angela~Yoonseo Park, and Jungyeul Park.
\newblock {A Novel Alignment-based Approach for PARSEVAL Measures}.
\newblock \emph{Computational Linguistics}, 50\penalty0 (3):\penalty0 1181--1190, 9 2024.
\newblock URL \url{https://aclanthology.org/2024.cl-3.10}.

\bibitem[Kim and Yang(2003)]{kim-yang-2003}
Jong-Bok Kim and Jaehyung Yang.
\newblock {K}orean phrase structure grammar and its implementations into the {LKB} system.
\newblock In Dong~Hong Ji and Kim~Teng Lua, editors, \emph{Proceedings of the 17th Pacific Asia Conference on Language, Information and Computation}, pages 88--97, Sentosa, Singapore, October 2003. COLIPS PUBLICATIONS.
\newblock \doi{http://hdl.handle.net/2065/12308}.
\newblock URL \url{https://aclanthology.org/Y03-1010/}.

\bibitem[Kim and Yang(2004)]{kim-yang-2004}
Jong-Bok Kim and Jaehyung Yang.
\newblock Projections from morphology to syntax in the {K}orean resource grammar: Implementing typed feature structures.
\newblock In Alexander Gelbukh, editor, \emph{Computational Linguistics and Intelligent Text Processing}, pages 14--25, Berlin, Heidelberg, 2004. Springer Berlin Heidelberg.
\newblock ISBN 978-3-540-24630-5.

\bibitem[Kim et~al.(2024)Kim, Chen, Jo, Lim, Park, and Park]{kim-etal-2024-kud}
Kyuwon Kim, Yige Chen, Eunkyul~Leah Jo, KyungTae Lim, Jungyeul Park, and Chulwoo Park.
\newblock {K-UD: Revising Korean Universal Dependencies Guidelines}.
\newblock \emph{arXiv}, pages 1--6, 2024.
\newblock URL \url{https://arxiv.org/abs/2412.00856}.

\bibitem[Kim and Park(2022)]{kim-park:2022}
Mija Kim and Jungyeul Park.
\newblock {A note on constituent parsing for Korean}.
\newblock \emph{Natural Language Engineering}, 28\penalty0 (2):\penalty0 199--222, 2022.
\newblock \doi{10.1017/S1351324920000479}.

\bibitem[Kitaev and Klein(2018)]{kitaev-klein-2018-constituency}
Nikita Kitaev and Dan Klein.
\newblock {Constituency Parsing with a Self-Attentive Encoder}.
\newblock In \emph{Proceedings of the 56th Annual Meeting of the Association for Computational Linguistics (Volume 1: Long Papers)}, pages 2675--2685, Melbourne, Australia, 7 2018. Association for Computational Linguistics.
\newblock URL \url{http://aclweb.org/anthology/P18-1249}.

\bibitem[Kitaev et~al.(2019)Kitaev, Cao, and Klein]{kitaev-cao-klein:2019:ACL}
Nikita Kitaev, Steven Cao, and Dan Klein.
\newblock {Multilingual Constituency Parsing with Self-Attention and Pre-Training}.
\newblock In \emph{Proceedings of the 57th Annual Meeting of the Association for Computational Linguistics}, pages 3499--3505, Florence, Italy, 7 2019. Association for Computational Linguistics.
\newblock URL \url{https://www.aclweb.org/anthology/P19-1340}.

\bibitem[Kuncoro et~al.(2017)Kuncoro, Ballesteros, Kong, Dyer, Neubig, and Smith]{kuncoro-etal-2017-recurrent}
Adhiguna Kuncoro, Miguel Ballesteros, Lingpeng Kong, Chris Dyer, Graham Neubig, and Noah~A. Smith.
\newblock {What Do Recurrent Neural Network Grammars Learn About Syntax?}
\newblock In Mirella Lapata, Phil Blunsom, and Alexander Koller, editors, \emph{Proceedings of the 15th Conference of the European Chapter of the Association for Computational Linguistics: Volume 1, Long Papers}, pages 1249--1258, Valencia, Spain, 4 2017. Association for Computational Linguistics.
\newblock URL \url{https://aclanthology.org/E17-1117}.

\bibitem[Liu and Zhang(2017)]{liu-zhang-2017-order}
Jiangming Liu and Yue Zhang.
\newblock {In-Order Transition-based Constituent Parsing}.
\newblock \emph{Transactions of the Association for Computational Linguistics}, 5:\penalty0 413--424, 2017.
\newblock \doi{10.1162/tacl_a_00070}.
\newblock URL \url{https://aclanthology.org/Q17-1029}.

\bibitem[Noh et~al.(2018)Noh, Han, Oh, and Kim]{noh-etal-2018-enhancing}
Youngbin Noh, Jiyoon Han, Tae~Hwan Oh, and Hansaem Kim.
\newblock {Enhancing Universal Dependencies for Korean}.
\newblock In \emph{Proceedings of the Second Workshop on Universal Dependencies (UDW 2018)}, pages 108--116, Brussels, Belgium, 11 2018. Association for Computational Linguistics.
\newblock \doi{10.18653/v1/W18-6013}.
\newblock URL \url{https://aclanthology.org/W18-6013}.

\bibitem[Park(2018)]{chulwoo-2019-for-developing}
Chulwoo Park.
\newblock {For Developing the Synthetic Perspective about the Concept of Word in Korean}.
\newblock \emph{HANGEUL}, 79\penalty0 (2):\penalty0 327--368, 2018.
\newblock ISSN 1225-0449.
\newblock \doi{https://doi.org/10.22557/HG.2018.06.79.2.327}.
\newblock URL \url{https://www.kci.go.kr/kciportal/ci/sereArticleSearch/ciSereArtiView.kci?sereArticleSearchBean.artiId=ART002356650}.

\bibitem[Park(2006)]{park:2006}
Jungyeul Park.
\newblock \emph{{Extraction automatique d'une grammaire d'arbres adjoints {\`{a}} partir d'un corpus arbor{\'{e}} pour le cor{\'{e}}en}}.
\newblock PhD thesis, Universit{\'{e}} Paris 7 - Denis Diderot, Paris, France, 2006.
\newblock URL \url{http://www.sudoc.fr/107995174}.

\bibitem[Park(2017)]{park:2017:Depling}
Jungyeul Park.
\newblock {Segmentation Granularity in Dependency Representations for Korean}.
\newblock In \emph{Proceedings of the Fourth International Conference on Dependency Linguistics (Depling 2017)}, pages 187--196, Pisa, Italy, 2017. Association for Computational Linguistics.
\newblock URL \url{http://aclweb.org/anthology/W/W17/W17-6522.pdf}.

\bibitem[Park and Kim(2023)]{park-kim-2023-role}
Jungyeul Park and Mija Kim.
\newblock {A role of functional morphemes in Korean categorial grammars}.
\newblock \emph{Korean Linguistics}, 19\penalty0 (1):\penalty0 1--30, 2023.
\newblock \doi{10.1075/kl.22003.par}.
\newblock URL \url{https://doi.org/10.1075/kl.22003.par}.

\bibitem[Park and Kim(2024)]{park-kim-2024-word}
Jungyeul Park and Mija Kim.
\newblock {Word segmentation granularity in Korean}.
\newblock \emph{Korean Linguistics}, 20\penalty0 (1):\penalty0 83--113, 2024.
\newblock URL \url{https://benjamins.com/catalog/kl.00008.par}.

\bibitem[Park and Park(2026)]{park-park-2026-constituency}
Jungyeul Park and Chulwoo Park.
\newblock {Constituency Structure over Eojeol in Korean Treebanks}, 2026.
\newblock URL \url{https://arxiv.org/abs/2512.22487}.

\bibitem[Park and Tyers(2019)]{park-tyers:2019:LAW}
Jungyeul Park and Francis Tyers.
\newblock {A New Annotation Scheme for the Sejong Part-of-speech Tagged Corpus}.
\newblock In \emph{Proceedings of the 13th Linguistic Annotation Workshop}, pages 195--202, Florence, Italy, 8 2019. Association for Computational Linguistics.
\newblock URL \url{https://www.aclweb.org/anthology/W19-4022}.

\bibitem[Park et~al.(2014)Park, Nam, Kim, Hahm, Hwang, and Choi]{park-EtAl:2014:ISWCPOSTER}
Jungyeul Park, Sejin Nam, Youngsik Kim, Younggyun Hahm, Dosam Hwang, and Key-Sun Choi.
\newblock {Frame-Semantic Web : a Case Study for Korean}.
\newblock In \emph{ISWC-PD'14: Proceedings of the 2014 International Conference on Posters {\&} Demonstrations Track - Volume 1272}, pages 257--260, Riva del Garda, Italy, 10 2014. International Semantic Web Conference.
\newblock URL \url{https://dl.acm.org/doi/10.5555/2878453.2878518}.

\bibitem[Park et~al.(2016)Park, Hong, and Cha]{park-hong-cha:2016:PACLIC}
Jungyeul Park, Jeen-Pyo Hong, and Jeong-Won Cha.
\newblock {Korean Language Resources for Everyone}.
\newblock In \emph{Proceedings of the 30th Pacific Asia Conference on Language, Information and Computation: Oral Papers (PACLIC 30)}, pages 49--58, Seoul, Korea, 2016. Pacific Asia Conference on Language, Information and Computation.
\newblock URL \url{http://aclweb.org/anthology/Y/Y16/Y16-2002.pdf}.

\bibitem[Park et~al.(2021)Park, Moon, Kim, Cho, Han, Park, Song, Kim, Song, Oh, Lee, Oh, Lyu, Jeong, Lee, Seo, Lee, Kim, Lee, Jang, Do, Kim, Lim, Lee, Park, Shin, Kim, Park, Oh, Ha, and Cho]{park-etal-2021-klue}
Sungjoon Park, Jihyung Moon, Sungdong Kim, Won~Ik Cho, Ji~Yoon Han, Jangwon Park, Chisung Song, Junseong Kim, Youngsook Song, Taehwan Oh, Joohong Lee, Juhyun Oh, Sungwon Lyu, Younghoon Jeong, Inkwon Lee, Sangwoo Seo, Dongjun Lee, Hyunwoo Kim, Myeonghwa Lee, Seongbo Jang, Seungwon Do, Sunkyoung Kim, Kyungtae Lim, Jongwon Lee, Kyumin Park, Jamin Shin, Seonghyun Kim, Lucy Park, Alice Oh, Jung-Woo Ha, and Kyunghyun Cho.
\newblock {KLUE: Korean Language Understanding Evaluation}.
\newblock In Joaquin Vanschoren and Serena Yeung, editors, \emph{Proceedings of the Neural Information Processing Systems Track on Datasets and Benchmarks}, volume~1, pages 1--25. Curran, 2021.

\bibitem[Petrov et~al.(2012)Petrov, Das, and McDonald]{petrov-das-mcdonald:2012:LREC}
Slav Petrov, Dipanjan Das, and Ryan McDonald.
\newblock {A Universal Part-of-Speech Tagset}.
\newblock In \emph{Proceedings of the Eighth International Conference on Language Resources and Evaluation (LREC-2012)}, pages 2089--2096, Istanbul, Turkey, 2012. European Language Resources Association (ELRA).
\newblock ISBN 978-2-9517408-7-7.

\bibitem[Post and Gildea(2009)]{post-gildea-2009-bayesian}
Matt Post and Daniel Gildea.
\newblock {Bayesian Learning of a Tree Substitution Grammar}.
\newblock In \emph{Proceedings of the ACL-IJCNLP 2009 Conference Short Papers}, pages 45--48, Suntec, Singapore, 2009. Association for Computational Linguistics.
\newblock URL \url{http://www.aclweb.org/anthology/P/P09/P09-2012}.

\bibitem[Sarkar(2002)]{sarkar:2002}
Anoop Sarkar.
\newblock \emph{{Combining labeled and unlabeled data in statistical natural language parsing}}.
\newblock PhD thesis, University of Pennsylvania, Philadelphia, Pennsylvania, USA, 2002.

\bibitem[Sarkar and Han(2002)]{sarkar-han:2002:TAG}
Anoop Sarkar and Chung-Hye Han.
\newblock {Statistical Morphological Tagging and Parsing of Korean with an LTAG Grammar}.
\newblock In \emph{Proceedings of 6th International Workshop on Tree Adjoining Grammars and Related Frameworks (TAG+6)}, pages 48--56, Venice, Italy, 2002.

\bibitem[Seddah et~al.(2013)Seddah, Tsarfaty, K{\"{u}}bler, Candito, Choi, Farkas, Foster, Goenaga, Gojenola~Galletebeitia, Goldberg, Green, Habash, Kuhlmann, Maier, Nivre, Przepi{\'{o}}rkowski, Roth, Seeker, Versley, Vincze, Woli{\'{n}}ski, Wr{\'{o}}blewska, and de~la Clergerie]{seddah-EtAl:2013:SPMRL}
Djamé Seddah, Reut Tsarfaty, Sandra K{\"{u}}bler, Marie Candito, Jinho~D. Choi, Richárd Farkas, Jennifer Foster, Iakes Goenaga, Koldo Gojenola~Galletebeitia, Yoav Goldberg, Spence Green, Nizar Habash, Marco Kuhlmann, Wolfgang Maier, Joakim Nivre, Adam Przepi{\'{o}}rkowski, Ryan Roth, Wolfgang Seeker, Yannick Versley, Veronika Vincze, Marcin Woli{\'{n}}ski, Alina Wr{\'{o}}blewska, and Eric~Villemonte de~la Clergerie.
\newblock {Overview of the SPMRL 2013 Shared Task: A Cross-Framework Evaluation of Parsing Morphologically Rich Languages}.
\newblock In \emph{Proceedings of the Fourth Workshop on Statistical Parsing of Morphologically-Rich Languages}, pages 146--182, Seattle, Washington, USA, 10 2013. Association for Computational Linguistics.
\newblock URL \url{http://www.aclweb.org/anthology/W13-4917}.

\bibitem[Seddah et~al.(2014)Seddah, K{\"{u}}bler, and Tsarfaty]{seddah-kubler-tsarfaty:2014}
Djamé Seddah, Sandra K{\"{u}}bler, and Reut Tsarfaty.
\newblock {Introducing the SPMRL 2014 Shared Task on Parsing Morphologically-rich Languages}.
\newblock In \emph{Proceedings of the First Joint Workshop on Statistical Parsing of Morphologically Rich Languages and Syntactic Analysis of Non-Canonical Languages}, pages 103--109, Dublin, Ireland, 8 2014. Dublin City University.
\newblock URL \url{https://www.aclweb.org/anthology/W14-6111}.

\bibitem[Seo et~al.(2019)Seo, Kim, Sung, and Yoo]{seo-etal-2019-ud}
Saetbyol Seo, Myeong-ju Kim, YeonSook Sung, and Seong~Hee Yoo.
\newblock {A Proposal on Universal Dependencies (v.2) Annotation for Korean}.
\newblock \emph{Language and Information}, 23\penalty0 (1):\penalty0 91--122, 2019.
\newblock URL \url{https://doi.org/10.29403/LI.23.1.5}.

\bibitem[Vinyals et~al.(2015)Vinyals, Kaiser, Koo, Petrov, Sutskever, and Hinton]{vinyals-etal-2015-grammar}
Oriol Vinyals, Lukasz Kaiser, Terry Koo, Slav Petrov, Ilya Sutskever, and Geoffrey~E. Hinton.
\newblock {Grammar as a Foreign Language}.
\newblock In C.~Cortes, N.~D. Lawrence, D.~D. Lee, M.~Sugiyama, and R.~Garnett, editors, \emph{Advances in Neural Information Processing Systems 28}, pages 2773--2781. Curran Associates, Inc., 2015.
\newblock URL \url{http://papers.nips.cc/paper/5635-grammar-as-a-foreign-language.pdf}.

\bibitem[Xia et~al.(2000)Xia, Han, Palmer, and Joshi]{xia-EtAl:2000:CLPW}
Fei Xia, Chunghye Han, Martha Palmer, and Aravind Joshi.
\newblock {Comparing Lexicalized Treebank Grammars Extracted from Chinese, Korean, and English Corpora}.
\newblock In \emph{Second Chinese Language Processing Workshop}, pages 52--59, Hong Kong, China, 2000. Association for Computational Linguistics.
\newblock \doi{10.3115/1117769.1117778}.
\newblock URL \url{http://www.aclweb.org/anthology/W00-1208}.

\bibitem[Yu~Cho and Sells(1995)]{cho-sells-1995}
Young-mee Yu~Cho and Peter Sells.
\newblock A lexical account of inflectional suffixes in {K}orean.
\newblock \emph{Journal of East Asian Linguistics}, 4\penalty0 (2):\penalty0 119--174, 1995.

\end{thebibliography}

\end{document}